%% file: main.tex
\documentclass[]{fairmeta}

\title{\vspace{-1pt}KnockGS: Interaction-Grounded Calibration of Physical Gaussian Representations}

\author{
Chenchen Ge$^{1,2,*}$,
Hanwen Shen$^{3,*}$,
Bowen Jing$^{1}$,
Jiyuan Cai$^{7}$,
Xiaofeng Wang$^{4,8}$,
Hongsen Lei$^{9}$,
Weitao Zhou$^{4,5}$,
Dandan Zhang$^{6}$,
Haibao Yu$^{1,10,\dagger}$
}

\affiliation[]{
\text{$^{1}$Tuojing Intelligence,
$^{2}$Southeast University,
$^{3}$Stevens Institute of Technology,
$^{4}$Tsinghua University,} \\
\text{ $^{5}$Simple AI,
$^{6}$Imperial College London,
$^{7}$Shanghai Jiao Tong University,
$^{8}$GigaAI,
$^{9}$Sun Yat-sen University,}\\
\text{$^{10}$The University of Hong Kong}
}

\contribution[ *]{Equal contribution}
\contribution[\dagger]{Corresponding author}

\input{math_commands.tex}

\usepackage{graphicx}
\usepackage[table]{xcolor}
\usepackage{colortbl}

\usepackage{booktabs}
\usepackage{tabularx}
\usepackage{array}
\usepackage{multirow}
\usepackage{diagbox}
\usepackage{hhline}
\usepackage{longtable}
\usepackage{makecell}
\usepackage{siunitx}
\usepackage{adjustbox}
\usepackage{wrapfig}
\usepackage{caption}   
\definecolor{oursgray}{RGB}{242,242,242}
\newcolumntype{C}{>{\centering\arraybackslash}X}

\usepackage{amsmath,amsfonts,amssymb}
\usepackage{bm}
\usepackage{nicefrac}

\usepackage{enumitem}
\setlist[itemize]{leftmargin=*}

\usepackage{caption}
\crefname{figure}{Fig.}{Figs.}
\crefname{table}{Tab.}{Tabs.}

\usepackage{xspace}
\usepackage{calc}
\usepackage{etoolbox}
\usepackage{pifont}
\usepackage{fancyvrb}
\usepackage{tikz}
\usetikzlibrary{positioning,arrows.meta}

\usepackage{titletoc}

\titlecontents{section}
[1.5em] %
{\addvspace{-0.5pt}} %
{\bfseries\contentslabel{2.3em}} %
{\hspace*{-2.3em}\bfseries} %
{\bfseries\titlerule*[.5pc]{.}\contentspage} %
\titlecontents{subsection}
[3.8em] %
{\addvspace{-2.2pt}} %
{\contentslabel{2.3em}}
{\hspace*{-2.3em}}
{\titlerule*[.5pc]{.}\contentspage}

\abstract{

Physics-integrated 3D Gaussian representations now allow reconstructed deformable objects to be simulated and rendered under explicit material models. Existing pipelines, however, assume that material parameters are known or manually specified, limiting their applicability when these parameters must be inferred from observed object dynamics. We propose \textbf{KnockGS}, an interaction-response PhysicalGS framework that estimates the elasticity and density scales of a 3D Gaussian object from its dynamics under a known applied force. Rather than treating physical simulation only as a forward process, we turn the force-induced response into a calibration signal: temporal response features are extracted from the observed dynamics, the two material scales are estimated from those features, and the estimate is then frozen and written back into the same simulator so that it can be tested on an interaction it was never fitted to.

We evaluate the framework on both parameter recovery and response-level fidelity. The estimated scales are compared against hidden ground truth, and the re-simulated object is measured against the target using 3D particle trajectories, response-curve statistics, and rendered-frame quality. Across five held-out material targets, our method recovers the scales substantially more accurately than response retrieval, global regression, or a fixed default material, and the frozen estimate remains predictive under interactions that differ in direction and in magnitude. Interaction response therefore carries enough information to calibrate material scales in physically grounded 3D Gaussian representations. Our study is a first step toward interactive PhysicalGS systems that calibrate a Gaussian asset whose rendered appearance and simulated response are consistent.

	\vspace{-10pt}
}
\metadata[Code]{
\url{https://github.com/TuojingAI/KnockGS}
}

\definecolor{lightgray}{rgb}{0.95, 0.95, 0.95}

\definecolor{baselinecolor}{gray}{.9}

\newcommand{\maybeinclude}[3]{%
  \IfFileExists{#1}{\includegraphics[width=#2]{#1}}{%
    \fbox{\parbox[c][#3][c]{0.85\linewidth}{\centering \scriptsize\texttt{\detokenize{#1}}\\ (placeholder)}}}}
\newcommand{\cmark}{\ding{51}}
\newcommand{\xmark}{\ding{55}}
\newcommand{\pmark}{(\ding{51})}

\begin{document}
\maketitle

\input{sec/1_intro}

\input{sec/2_related_work}
\input{sec/3_method}
\input{sec/4._experiments}

\input{sec/5._conclusion}

\bibliography{main}
\bibliographystyle{bibstyle}

\newpage
\beginappendix
\input{sec/6._appendix}

\end{document}

%% file: math_commands.tex
\usepackage{amsmath,amsfonts,bm}
\usepackage{xcolor}

\def\eqref#1{equation~\ref{#1}}

\def\1{\bm{1}}

\DeclareMathAlphabet{\mathsfit}{\encodingdefault}{\sfdefault}{m}{sl}
\SetMathAlphabet{\mathsfit}{bold}{\encodingdefault}{\sfdefault}{bx}{n}

%% file: sec/1_intro.tex
\section{Introduction}
\label{sec:intro}


\begin{figure*}[!t]
  \centering
  \includegraphics[width=0.96\textwidth,trim=0 60bp 0 0,clip]{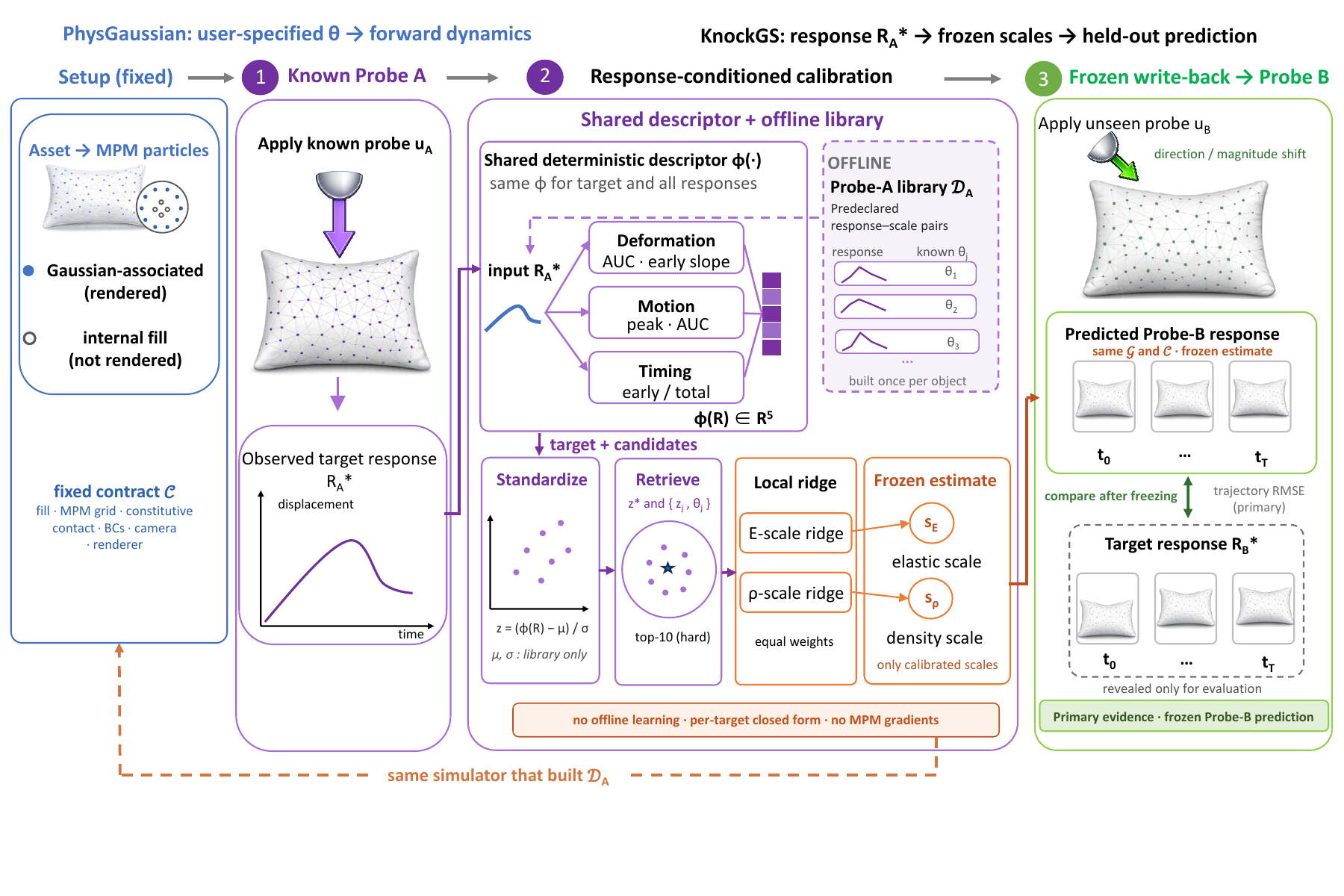}
  \caption{Overview of KnockGS and its frozen-prediction protocol. The unnumbered setup fixes the simulatable Gaussian asset $\mathcal{G}$ and simulator contract $\mathcal{C}$. (1) A known Probe~A, $u_A$, produces the observed target response $R_A^\star$. (2) The same deterministic five-dimensional descriptor $\phi$ encodes that response and all responses in the precomputed object-specific library $\mathcal{D}_A$; library-only standardization, hard top-10 retrieval, and an equal-weight local ridge fit yield continuous elasticity and density scales $(s_E,s_\rho)$. (3) The estimate is frozen and written back into the same simulator to predict $\widehat{R}_B$ under the held-out Probe~B, $u_B$. The target $R_B^\star$ is revealed only after freezing, and trajectory RMSE is the primary evidence. Purple dashed paths denote once-per-object offline construction and reuse; the orange dashed return identifies the same simulator contract that built $\mathcal{D}_A$.}
  \label{fig:pipeline}
\end{figure*}

Three-dimensional visual reconstruction can now recover the geometry and appearance of real objects with high fidelity, yet it cannot answer a basic question: what happens to the object when a force is applied to it? When PhysicalGS is to serve as foundational digital asset, i.e., the underlying representation in robotic manipulation, then its simulated physical behavior and its rendered appearance must agree. Robotic manipulation, Physical AI, digital twins, and interactive simulation all require this predictive capability, motivating a step beyond visual reconstruction toward \emph{3D physical reconstruction}. On the visual side, neural radiance fields and their accelerations have reshaped novel-view synthesis~\cite{mildenhallnerf,instantngp,mipnerf360}, while 3D Gaussian Splatting (3DGS) provides an explicit representation with real-time rendering~\cite{kerbl3dgs}. Physics-integrated extensions such as PhysGaussian~\cite{xiephysgaussian} make Gaussian primitives simulatable by associating them with material points evolved by the Material Point Method (MPM)~\cite{sulskympm,stomakhinsnow,jiangmpm,humlsmpm}. Separate visual and physical representations must be converted into one another, kept in correspondence, and rendered consistently; a physics-integrated Gaussian asset avoids all three by unifying appearance and mechanical state. However, such pipelines are predominantly \emph{forward}: the material model and its parameters must be specified before simulation can predict the resulting dynamics.



Gaussians are a natural carrier of physical state. Their particle-like structure maps directly onto MPM dynamics without requiring an explicit mesh, while retaining appearance and mechanical state within a unified representation. Established 3DGS reconstruction pipelines can therefore provide the geometric and visual basis for simulatable assets; what remains unresolved is how to determine their physical parameters. Specifically, visual reconstruction alone does not determine elasticity, density, damping, friction, or internal structure. \textbf{Two objects with very different mass and stiffness can fall and appear similarly under gravity.} Objects with nearly identical appearance can exhibit substantially different responses to the same physical interaction. Static appearance and passive observation are therefore fundamentally limited in disambiguating such objects, whereas an \emph{actively applied, known} interaction together with its observed response provides direct physical evidence about the object. Appearance-based physical-property methods~\cite{gaussianproperty,pugs,physgs,physgm} and video-based inverse or dynamics-learning methods~\cite{pacnerf,gic,springgaus,phystwin,projo4d,reconphys} provide useful alternatives, but they do not directly answer whether a parameter estimate obtained from one \emph{known controlled interaction} predicts a different, unseen interaction. 

The target research problem is interaction-grounded physical representation: from an initial observation $O_0$, a known interaction $u_A$, and a physically observable response $Y_A$, infer a representation $z_{\mathrm{phys}}$ that predicts the response $Y_B$ to a disjoint interaction $u_B$. The decisive criterion is therefore not parameter proximity alone, but whether the frozen representation predicts motion, deformation, contact, and ultimately visual response under $u_B$. Real systems would instantiate $Y_A$ with RGB/RGB-D video, surface tracks, silhouettes, or force/tactile measurements.

This paper studies a controlled version of interaction-grounded physical inference (Fig.~\ref{fig:pipeline}): rather than supplying material parameters for forward simulation, can the physical parameters inferred from one known interaction predict the response to a different, unseen interaction? The unnumbered setup in Fig.~\ref{fig:pipeline} fixes the simulatable Gaussian asset (following PhysGaussian~\cite{xiephysgaussian}), MPM solver, particle fill, grid, constitutive family, boundary conditions, and rendering. Within this fixed contract, we estimate only two dimensionless material scales, a Young's modulus scale and a density scale $\boldsymbol{\theta}=(s_E,s_\rho)$. The observed response is simulator-exported and therefore \emph{privileged}, with both parameter and response ground truth known.

We separate calibration from prediction using two interactions. Probe A is a known controlled interaction used to estimate $\boldsymbol{\theta}$; Probe B differs in direction, magnitude, or both and is never used during estimation. The estimated parameters are frozen before Probe B is applied. Thus, Probe A measures whether the method can recover a physically useful state, while Probe B tests whether that state predicts beyond the interaction it was fitted to.

KnockGS, an interaction-response PhysicalGS performs this calibration with an object-specific library of 54 Probe-A responses, precomputed once under the same simulator contract. A shared deterministic five-dimensional descriptor encodes both the target and every candidate response; after library-only standardization and hard top-10 retrieval, an equal-weight local ridge fit predicts continuous $(s_E,s_\rho)$ values without differentiating through MPM. The estimate is then frozen and written back into the same simulator. We use held-out Probe-B trajectory RMSE as the primary evidence, with rendered PSNR/SSIM and response-curve statistics as complementary measures.


On held-out Pillow targets, the local estimator outperforms response-nearest, response-kNN, and global ridge given identical evidence, and remains predictive under direction- and magnitude-shifted Probe B. The protocol also succeeds on Ficus and Vasedeck. At the same time, observation degradation, fill mismatch, grid mismatch, and cross-object transfer expose clear failure boundaries. These results position the method not as a universal material representation, but as an object- and discretization-conditioned calibration mechanism whose validity is tested by cross-interaction prediction.

Our contributions are:

\begin{enumerate}
  \item We formulate material-scale inference for physics-integrated Gaussians as a Probe-A-to-Probe-B problem, where the estimate from one known interaction is frozen and judged by its prediction of an unseen one, unlike forward-only PhysGaussian pipelines and passive-video inverse methods.

  \item We propose a library-based local estimator that yields continuous $(s_E,s_\rho)$ without per-target simulation or gradients through MPM. Under the same evidence it reduces joint scale error from 2.4\% (KNN, global ridge) to 1.1\% and lowers held-out Probe-B trajectory error by about $3\times$, while also outperforming per-target CMA-ES video optimization.

  \item We show that a known probe resolves the stiffness-to-mass ambiguity inherent to passive observation: our estimates track the full 162\% spread of the held-out targets, whereas exceed the performance of baseline models. We further give an explicit account of where the method fails.
\end{enumerate}

%% file: sec/2_related_work.tex
\section{Related Work}
\label{sec:related-work}

\begin{table*}[!t]
\centering
\caption{Comparison with representative related work in the PhysGaussian ecosystem.}
\label{tab:related_work}
\scriptsize
\setlength{\tabcolsep}{2.8pt}
\renewcommand{\arraystretch}{1.08}

\begin{tabularx}{\textwidth}{
>{\raggedright\arraybackslash}p{2.25cm}
>{\raggedright\arraybackslash}X
*{5}{>{\centering\arraybackslash}p{1.28cm}}
}
\toprule
Category
& Representative works
& GS / particle repr.
& Material estimation
& Interaction-Response probing
& 3D response features
& Re-simulation fidelity \\
\midrule
PhysGaussian base
& PhysGaussian~\cite{xiephysgaussian}
& \cmark & \xmark & \xmark & \xmark & \xmark \\

Solver / forward enhancement
& i-PhysGaussian~\cite{iphysgaussian}, FastPhysGS~\cite{fastphysgs}, GaussianFluent~\cite{gaussianfluent}
& \cmark & \xmark & \xmark & \xmark & \xmark \\

Multi-material / scene physics
& OmniPhysGS~\cite{linomniphysgs}, PhysSplat~\cite{physsplat}
& \cmark & \pmark & \xmark & \xmark & \xmark \\

Visual / VLM property assignment
& GaussianProperty~\cite{gaussianproperty}, PUGS~\cite{pugs}, PhysGS~\cite{physgs}, PhysGM~\cite{physgm}
& \cmark & \cmark & \xmark & \xmark & \xmark \\

Inverse problem / digital twin
& GIC~\cite{gic}, Spring-Gaus~\cite{springgaus}, PhysTwin~\cite{phystwin}, ReconPhys~\cite{reconphys}, PAC-NeRF~\cite{pacnerf}
& \pmark & \cmark & \xmark & \pmark & \pmark \\

Dynamics / world models
& NGFF~\cite{ngff}, Dynamic 3D Gaussian tracking~\cite{luitendynamic3dgs}, DiffWind~\cite{diffwind}
& \pmark & \xmark & \xmark & \pmark & \xmark \\

Robot online adaptation
& AdaptiGraph~\cite{adaptigraph}, ManiGaussian~\cite{manigaussian}, SplatSim~\cite{splatsim}
& \pmark & \pmark & \xmark & \xmark & \xmark \\
\midrule
\rowcolor{oursgray}
\textbf{Response-conditioned calibration}
& \textbf{KnockGS (ours)}
& \cmark & \cmark & \cmark & \cmark & \cmark \\
\bottomrule
\end{tabularx}

\vspace{0.3em}
\begin{minipage}{0.96\textwidth}
\scriptsize
\textit{Note:} \cmark: satisfied; \xmark: not addressed; \pmark: partially satisfied
\end{minipage}
\end{table*}

\subsection{Physics-Integrated Gaussian Representations}
PhysGaussian~\cite{xiephysgaussian} demonstrated that 3D Gaussians~\cite{kerbl3dgs} can serve as a unified representation for physics-based simulation and rendering, extending neural scene representations~\cite{mildenhallnerf} with MPM-based dynamics~\cite{sulskympm,jiangmpm}. However, its breadth does not imply that follow-up works address effective material-scale calibration: in our manual survey of PhysGaussian-related follow-up work, forward simulation, interactive editing, dynamic 4D reconstruction, general reconstruction, and generative content account for the overwhelming majority, while only a small fraction directly addresses physical property estimation, inverse problems, or calibration. We target this under-explored setting, with a distinct evidence chain: controlled interaction-response probing, particle-level 3D response features, response-space calibration, and re-simulation fidelity as the primary evaluation. 

\subsection{Physical System Identification from Visual Dynamics}

The ecosystem can be organized into three adjacent routes. Forward PhysicalGS methods assume material parameters and improve simulation or rendering~\cite{xiephysgaussian,linomniphysgs,iphysgaussian,fastphysgs}. Visual-prior approaches infer or assign attributes from static appearance~\cite{gaussianproperty,pugs,physgs,physgm}. Passive-video inverse methods use observed dynamics, often through differentiable optimization or learned prediction~\cite{pacnerf,gic,springgaus,reconphys}. PhysTwin~\cite{phystwin} explicitly takes sparse videos of deformable objects under interaction and evaluates simulation under novel interactions. Our controlled route differs in the information source and validation axis: the excitation is known, repeatable, and actively specified, and parameters estimated from Probe A are frozen before a different Probe B is evaluated. Table~\ref{tab:related_work} summarizes these distinctions.



%% file: sec/3_method.tex
\section{Method}
\label{sec:method}

Given a fixed PhysicalGS asset $\mathcal{G}$ and simulator contract $\mathcal{C}$, our method uses the observed response to a known \emph{Calibration Probe A} $u_A$ to estimate two dimensionless material scales, $\boldsymbol{\theta}=(s_E,s_\rho)$. The estimate is then frozen, written back into the same PhysicalGS simulator, and used to predict the response to a disjoint \emph{Held-out Prediction Probe B} $u_B$. The asset, particle fill, MPM solver and grid, constitutive family, boundary conditions, camera, and renderer are built on the PhysGaussian backbone~\cite{xiephysgaussian} and remain fixed within $\mathcal{C}$; only the two scales are calibrated. In the terminology of Sec.~\ref{sec:intro}, the fixed asset $\mathcal{G}$ represents the initial observation $O_0$, the target response $R_A^\star$ instantiates $Y_A$, the calibrated asset $(\mathcal{G},\widehat{\boldsymbol{\theta}})$ under $\mathcal{C}$ instantiates $z_{\mathrm{phys}}$, and $\widehat{R}_B$ is the predicted instantiation of $Y_B$.


The design is motivated by the high cost of repeated MPM simulation. Rather than differentiating through the simulator or repeatedly optimizing each target, we precompute a small response library that can be reused for calibration. Sec.~\ref{sec:method:library} precomputes an object-specific response library and maps both library responses and target responses to the same deterministic five-dimensional descriptor. This enables response comparison in a compact feature space and allows a target to be matched against precomputed simulations. Figure~\ref{fig:pipeline} therefore separates the unnumbered fixed setup from a three-stage per-target closed loop: (1) apply the known Probe A and record the target response; (2) calibrate continuous material scales using the shared descriptor and a local closed-form fit; and (3) freeze the estimate, write it back into the same simulator, and predict the held-out Probe B.

\subsection{Problem Setting and Controlled Scope}
\label{sec:method:formulation}

We study effective material-scale calibration within a declared simulation
contract. A PhysicalGS asset $\mathcal{G}$ contains renderable Gaussian kernels and a simulator particle set
\begin{equation}
X^0=X^0_{\mathrm{GS}}\cup X^0_{\mathrm{fill}},
\end{equation}
where Gaussian centers are mapped to Gaussian-associated MPM particles and additional particles fill the interior. The contract $\mathcal{C}$ fixes the asset, coordinate transform, fill construction, MPM grid, constitutive family, contact model, boundary conditions, damping, friction, Poisson ratio, camera, and renderer. The only unknown quantities are
\begin{equation}
\boldsymbol{\theta}=(s_E,s_\rho)\in\Theta,
\end{equation}
where, $s_E$ and $s_\rho$ are effective calibration scales defined relative
to the fixed object-specific simulator template, rather than direct estimates
of intrinsic material constants.

An interaction protocol $u$ specifies the force vector, contact region, start time, and duration under the fixed support and boundary configuration. The simulator $\mathcal{S}$ produces a rollout under $u$. At predeclared export times $t\in\mathcal{T}_{\mathrm{out}}$ (31 frames in the main benchmark, including the initial state), the main observation operator exports particle positions,
\begin{equation}
R_u(\boldsymbol{\theta})=
\mathcal{O}_{\mathrm{part}}\!\left(
\mathcal{S}(\mathcal{G},\mathcal{C},\boldsymbol{\theta},u)
\right)
=\{\mathbf{x}^{t}_{i}\}_{i=1,t\in\mathcal{T}_{\mathrm{out}}}^{N}.
\label{eq:particle_response}
\end{equation}
The target response to Calibration Probe A is $R_A^\star=R_{u_A}(\boldsymbol{\theta}^\star)$, while $\boldsymbol{\theta}^\star$ remains hidden from every deployable estimator. Given the object-specific Probe-A library $\mathcal{D}_A$, the instantiated objective is
\begin{equation}
\widehat{\boldsymbol{\theta}}
=h(R_A^\star;\mathcal{D}_A),
\qquad
\widehat{R}_B=R_{u_B}(\widehat{\boldsymbol{\theta}}).
\label{eq:target_problem}
\end{equation}
The held-out target response $R_B^\star$ is used only to evaluate $\widehat{R}_B$ after calibration has been frozen. The main estimator therefore uses privileged simulator-exported state rather than captured real RGB/RGB-D; Sec.~\ref{sec:exp:observation} evaluates progressively less privileged synthetic observations.

\subsection{Object-Specific Response Library}
\label{sec:method:library}

\paragraph{Offline response bank.}
The response map $\boldsymbol{\theta}\mapsto R_A(\boldsymbol{\theta})$ is available only through simulation. We therefore evaluate it in advance on a fixed candidate set and reuse those responses for every target governed by the same $\mathcal{C}$ and $u_A$. For each asset, the resulting library is
\begin{equation}
\mathcal{D}_A=\{(R_A(\boldsymbol{\theta}_j),\boldsymbol{\theta}_j)\}_{j=1}^{J}
\end{equation}
, where $J$ is the number of candidates. The candidates form a predeclared, non-Cartesian set of parameter pairs rather than a regular grid. They provide finite coverage of the declared two-scale domain at a reusable offline cost. Appendix Sec.~\ref{app:probes} reports the exact support and target split, and Appendix Sec.~\ref{app:offline_sensitivity} measures how candidate count affects accuracy and coverage stability. All held-out targets are excluded from $\mathcal{D}_A$, and neither their hidden scales nor any Probe-B response contributes to library construction. We need a lightweight interpolation mechanism that converts a finite response library into continuous material estimates without additional simulator calls. Hard-neighborhood local ridge is the simplest mechanism we found that satisfies this requirement while outperforming retrieval and global regression under the same evidence. 

\paragraph{Shared response descriptor.}
Raw particle trajectories are high-dimensional and object-dependent, so the same deterministic map compresses every candidate and target response before comparison. Let $D_{\mathrm{bbox}}$ be the diagonal of the initial particle bounding box. We first compute normalized RMS displacement and frame-difference velocity curves,
\begin{equation}
r^t=\frac{1}{D_{\mathrm{bbox}}}
\sqrt{\frac{1}{N}\sum_i\|\mathbf{x}_i^t-\mathbf{x}_i^0\|_2^2},
\quad
v^t=\frac{1}{D_{\mathrm{bbox}}}
\sqrt{\frac{1}{N}\sum_i\|\mathbf{x}_i^t-\mathbf{x}_i^{t-1}\|_2^2}.
\label{eq:rms_curve}
\end{equation}
One five-dimensional descriptor then summarizes complementary response-amplitude and timing cues,
\begin{equation}
\phi(R)=\left[
\mathrm{AUC}(r),\; s_{\mathrm{early}}(r),\;
\max_t v^t,\; \mathrm{AUC}(v),\;
\frac{\sum_{t\leq t_e}v^t}{\sum_t v^t}
\right]^{\!\top},
\label{eq:descriptor}
\end{equation}
with $t_e=5$. Its entries measure cumulative deformation, early deformation rate, peak frame-to-frame motion, cumulative motion, and the fraction of motion concentrated early in the rollout, respectively. Every feature dimension is standardized using candidate-library statistics only. The descriptor has no learned encoder and is not assumed to make the two scales globally identifiable; it supplies compact local evidence for the calibration stage below.

\subsection{Response-Conditioned Continuous Calibration}
\label{sec:method:estimator}

Returning the nearest response can only select an existing candidate, so it cannot represent a target whose scales are absent from the finite library. A continuous interpolator is therefore required. Because a single linear map need not approximate the descriptor-to-scale relation across the full domain, we instead assume that it is better approximated within a small response-space neighborhood. A hard neighborhood provides this locality, while weak ridge regularization stabilizes the resulting few-sample fit. This design relaxes the quantization limitation of retrieval and produces a continuous estimate without a new MPM rollout or differentiation through the simulator. Sec.~\ref{sec:exp:main} evaluates the resulting local closed-form interpolation against response retrieval and global ridge under the same response library and descriptor evidence.

Let $\mathbf{z}_j$ and $\mathbf{z}^\star$ be the standardized descriptors of candidate $j$ and the target response. Euclidean distance is used only to select the hard neighborhood $\mathcal{N}_{10}$ containing the ten nearest candidates. Samples within that neighborhood have equal weight, and we solve
\begin{equation}
\min_{\mathbf{A},\mathbf{b}}
\sum_{j\in\mathcal{N}_{10}}
\|\mathbf{A}\mathbf{z}_j+\mathbf{b}-\boldsymbol{\theta}_j\|_2^2
+10^{-3}\|\mathbf{A}\|_F^2,
\qquad
\widehat{\boldsymbol{\theta}}=\mathbf{A}\mathbf{z}^\star+\mathbf{b}.
\label{eq:local_linear}
\end{equation}
The intercept is not regularized, and the two material scales are regressed independently. Predictions are clipped to the candidate-set bounds. Inverse-distance KNN is used only if fewer than three neighbors are available or a numerical exception prevents the ridge solve; the main estimator is not a distance-weighted ridge regression. The configuration $k=10$, $\alpha=10^{-3}$ is fixed before held-out evaluation. Probe B and all held-out target statistics remain unavailable during standardization, neighborhood retrieval, fitting, and configuration selection.

\subsection{Frozen Write-Back and Held-out Prediction}
\label{sec:method:validation}

Parameter proximity alone does not establish a useful PhysicalGS representation: an estimate may be numerically close to the hidden scales yet reproduce the wrong motion, or it may reconstruct Probe A without carrying information that transfers to a different interaction. We therefore close the loop in response space. Immediately after observing Calibration Probe A, $\widehat{\boldsymbol{\theta}}$ is frozen and written back into the same simulator that generated $\mathcal{D}_A$. We then distinguish calibration reconstruction from held-out prediction,
\begin{equation}
\widehat{R}_A=R_{u_A}(\widehat{\boldsymbol{\theta}}),
\qquad
\widehat{R}_B=R_{u_B}(\widehat{\boldsymbol{\theta}}).
\label{eq:ab_validation}
\end{equation}
Probe-A reconstruction is a diagnostic of consistency with the observed response. Held-out Prediction Probe B is the primary evidence: its protocol is predeclared, may change force direction, magnitude, duration, or contact location, and contributes no frames, trajectories, descriptors, or statistics to estimator design or fitting. Its target response $R_B^\star$ is revealed only for final comparison with $\widehat{R}_B$.

The primary dynamics comparison uses Gaussian-associated particles with stable identity,
\begin{equation}
\mathrm{RMSE}_{\mathrm{traj}}=
\sqrt{\frac{1}{|\mathcal{I}_{\mathrm{GS}}||\mathcal{T}_{\mathrm{out}}|}
\sum_{i\in\mathcal{I}_{\mathrm{GS}}}\sum_{t\in\mathcal{T}_{\mathrm{out}}}
\|\widehat{\mathbf{x}}_i^t-\mathbf{x}_i^{\star,t}\|_2^2}.
\label{eq:traj_rmse}
\end{equation}
Sec.~\ref{sec:experiments} complements this spatiotemporal metric with response-curve and rendered-frame measures. Accordingly, Probe-B evaluation measures cross-interaction prediction within the fixed object-specific simulator contract and should not be interpreted as arbitrary-interaction or real-material generalization.

%% file: sec/4._experiments.tex
\section{Experiments}
\label{sec:experiments}


This section tests the two claims made in Sec.~\ref{sec:method} and then maps their limits. The claims are that a local fit on a hard neighborhood extracts more from a response library than retrieval or global regression does (Sec.~\ref{sec:method:estimator}), and that the resulting frozen estimate predicts an excitation it was never fitted to (Sec.~\ref{sec:method:validation}). Sec.~\ref{sec:exp:setup} first fixes the experimental contract; the experiments then follow the order in which the claims can fail. We ask three questions: (1) under identical Probe-A evidence, does local calibration improve over deployable baselines; (2) after the estimate is frozen, does it predict held-out Probe B; and (3) can the object-specific loop be repeated across assets, and which observation, object, discretization, and identifiability boundaries limit it? Questions (1) and (2) test the claims directly, while question (3) separates procedural repeatability from the scope that Sec.~\ref{sec:exp:scope} states explicitly.

\subsection{Experimental Contract}
\label{sec:exp:setup}

\paragraph{Input and output.}
At estimation time a deployable method receives exactly three inputs: the frozen Gaussian asset and its simulation contract, the specification of the calibration probe $u_A$, and the response $R_A^\star$ that the hidden target produced under that probe. It outputs a single pair $\widehat{\boldsymbol{\theta}}=(\hat{s}_E,\hat{s}_\rho)$. During calibration, the estimator has no access to the target scales, any Probe-B observation, or aggregate statistics computed across held-out targets. The output is then consumed in one way only: it is written into the same simulator and re-simulated, first under $u_A$ and then under $u_B$, and all reported numbers are computed from those rollouts and their renders. The response used by the main estimator is simulator-exported particle state and is therefore \emph{privileged}; Sec.~\ref{sec:exp:observation} replaces it with progressively less privileged synthetic observations.

\paragraph{Baselines.}
All deployable in-domain methods receive no information beyond the same Probe-A evidence. Response-nearest, inverse-distance response KNN, global ridge, and our local ridge use the same predeclared object-specific response library and five-dimensional descriptor; fixed default ignores that evidence by construction. They differ only in how neighborhood evidence is converted into an estimate, which isolates the contribution of Sec.~\ref{sec:method:estimator}. KNN neighborhood size and global-ridge regularization are selected by candidate-only validation. Parameter-nearest reads the hidden target location and is reported only as an oracle diagnostic. Sec.~\ref{sec:exp:alternatives} separately evaluates direct video
optimization, PhysGM, and ReconPhys. Because these methods differ in both
input evidence and native parameterization, we report them as diagnostic
external comparisons rather than include them in the same-information
ranking.

\paragraph{Computing resources.}
The cost of the method is concentrated offline and paid once per object. Building a library requires 54 MPM rollouts, each covering $0.60$\,s of simulated time at a $10^{-4}$\,s substep and exporting 31 states; the Pillow asset carries $695{,}984$ particles on a background grid with $n_{\mathrm{grid}}=100$. Calibrating a new target adds no rollout: the estimator standardizes a five-dimensional vector, selects ten neighbors, and solves two ten-sample ridge systems in closed form — no training, no gradients through MPM, no learned weights. Evaluation is the dominant remaining cost, since every reported estimate is re-simulated under Probe A and each Probe B protocol. The CMA-ES baseline in Sec.~\ref{sec:exp:alternatives} is the one method whose cost scales with the number of targets: it spends 30 forward simulator calls per target per seed, i.e.\ 450 rollouts across five targets and three predeclared seeds. 


\paragraph{Evaluation metrics.}
We report the joint relative error
\begin{equation}
e_\theta=\frac{1}{2}\left(
\frac{|\hat{s}_E-s_E^\star|}{s_E^\star}+
\frac{|\hat{s}_\rho-s_\rho^\star|}{s_\rho^\star}
\right)\times100\%,
\end{equation}
and the Gaussian-associated-particle trajectory RMSE in Eq.~\eqref{eq:traj_rmse}. Curve RMSE, peak/AUC/final errors, foreground PSNR, and object-crop SSIM are secondary. Trajectories are compared at the same exported times with fixed particle correspondence. The in-domain estimators and MPM evaluations are deterministic under frozen seeds; Appendix Sec.~\ref{app:targetwise} therefore reports every held-out target and mean $\pm$ sample standard deviation across targets, rather than pseudo-replicating frames or particles. The stochastic CMA-ES baseline is additionally reported per target as mean $\pm$ standard deviation over its three predeclared seeds. Completion, leakage, missing-frame, NaN, and duplicate-output audits accompany the experiment packages.

\paragraph{Assets, candidates, and targets.}
The main benchmark uses a reconstructed pillow/sofa Gaussian asset and a candidate set formed by combining a broad sweep over the declared two-scale domain with a denser sweep that supplies local interpolation support. Removing duplicate parameter pairs yields 54 unique $(s_E,s_\rho)$ candidates. This set was fixed before any held-out evaluation; Appendix Sec.~\ref{app:offline_sensitivity} reports post-hoc sensitivity to smaller candidate subsets without retuning the main result. The five held-out scale pairs lie within the covered domain but do not
coincide with any library candidate, so the evaluation tests continuous scale
estimation rather than exact table lookup. The same five target locations are used for object-specific Ficus and Vasedeck experiments. Candidate and target responses are generated by the same PhysicalGS/MPM implementation. Their parameters are preset by the experimenter and hidden from deployable estimators; they are simulator-defined ground truth, not measured real-material properties. Appendix Tables~\ref{tab:object_contract}--\ref{tab:probe_values} give the exact split, material bases, timing, forces, contact regions, and particle populations.

\paragraph{Probe separation.}
The calibration interaction is Probe A (\texttt{standard\_x}). Pillow prediction uses Probe B with a new direction (\texttt{standard\_y}) and a larger magnitude (\texttt{strong\_x}); Vasedeck uses held-out \texttt{standard\_z}. Direction and magnitude are varied separately so that a failure to transfer can be attributed to one or the other. Fine-resolution duration and contact-shift diagnostics are reported quantitatively in Appendix Sec.~\ref{app:auxiliary_probes}. Probe B never participates in descriptor design, hyperparameter selection, neighborhood retrieval, or parameter estimation.

\subsection{Same-Information Calibration and Frozen Probe-B Prediction}
\label{sec:exp:main}

This experiment answers questions (1) and (2) together, and it is the only place where all deployable methods are guaranteed to see identical information. Parameter recovery tests whether the local fit extracts more from the library than its alternatives; frozen Probe-B prediction is the primary evidence that the estimate survives a change of excitation.

\begin{figure*}[!t]
\centering
\includegraphics[width=0.90\textwidth]{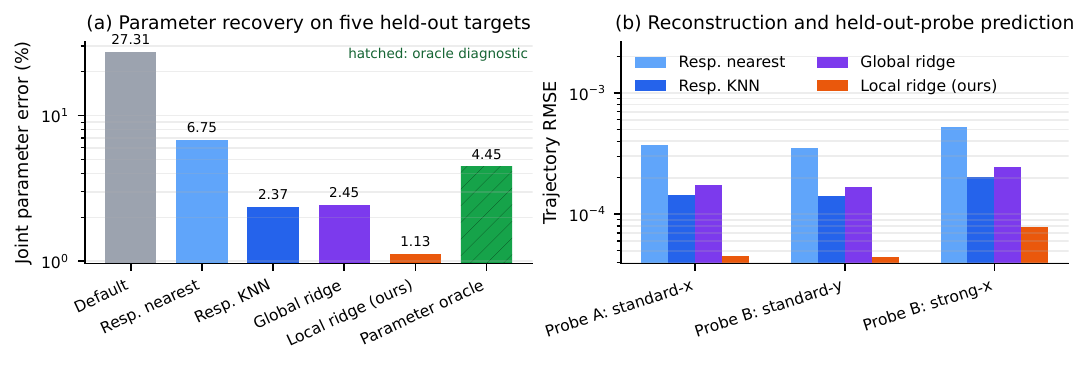}
\caption{Fair baseline comparison. (a) Local ridge gives the lowest mean joint error among deployable methods; the hatched oracle accesses hidden target parameters. (b) The primary prediction evidence: Probe-A reconstruction and held-out Probe-B trajectories are generated with the same frozen estimates.}
\label{fig:v59_main}
\end{figure*}

Table~\ref{tab:v59_main} and Fig.~\ref{fig:v59_main} give the central result. Local ridge achieves 1.13\% mean parameter error, improving over response KNN (2.37\%), global ridge (2.45\%), and response-nearest (6.75\%); this establishes better calibration under matched evidence. The primary evidence is the held-out prediction made by the same frozen estimate. On \texttt{standard\_y}, its Probe-B trajectory RMSE is $4.42\times10^{-5}$, versus $1.41\times10^{-4}$ for KNN and $1.69\times10^{-4}$ for global ridge. On \texttt{strong\_x}, it obtains $7.79\times10^{-5}$, versus $2.02\times10^{-4}$ and $2.47\times10^{-4}$. These results show that scales calibrated from Probe A remain predictive
under held-out changes in force direction and magnitude within the same
simulation contract.

\begin{table*}[!t]
\centering
\caption{Pillow benchmark with identical information for all deployable methods. Probe A is used for estimation; both Probe B protocols are unseen during estimation and tuning. Lower is better. Parameter-nearest accesses hidden target parameters and is an oracle diagnostic, not a deployable baseline.}
\label{tab:v59_main}
\footnotesize
\setlength{\tabcolsep}{3pt}
\begin{tabular}{lrrrr}
\toprule
Method & \shortstack{Joint parameter\\error (\%)} & \shortstack{Probe A\\traj. RMSE} & \shortstack{Probe B\\direction} & \shortstack{Probe B\\magnitude} \\
\midrule
Fixed default & 27.31 & $3.682{\times}10^{-3}$ & $3.502{\times}10^{-3}$ & $4.407{\times}10^{-3}$ \\
Response nearest & 6.75 & $3.722{\times}10^{-4}$ & $3.533{\times}10^{-4}$ & $5.188{\times}10^{-4}$ \\
Response KNN & 2.37 & $1.449{\times}10^{-4}$ & $1.412{\times}10^{-4}$ & $2.020{\times}10^{-4}$ \\
Global ridge & 2.45 & $1.748{\times}10^{-4}$ & $1.687{\times}10^{-4}$ & $2.466{\times}10^{-4}$ \\
\rowcolor{oursgray}\textbf{Local ridge (ours)} & \textbf{1.13} & $\mathbf{4.498{\times}10^{-5}}$ & $\mathbf{4.424{\times}10^{-5}}$ & $\mathbf{7.786{\times}10^{-5}}$ \\
\midrule
Parameter-nearest oracle & 4.45 & $6.283{\times}10^{-4}$ & $5.984{\times}10^{-4}$ & $7.144{\times}10^{-4}$ \\
\bottomrule
\end{tabular}
\end{table*}

\begin{figure*}[!t]
\centering
\includegraphics[width=0.94\textwidth]{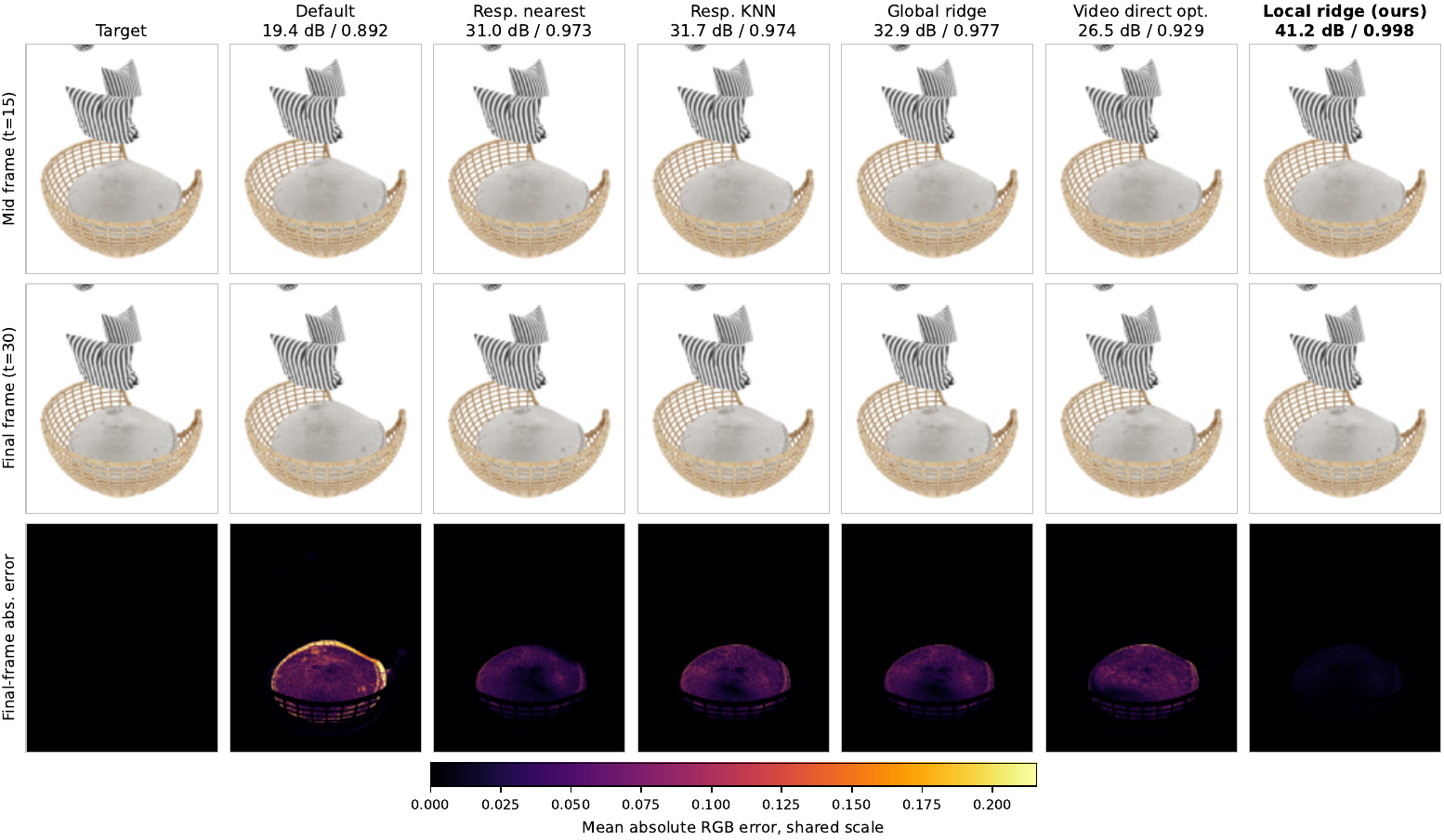}
\caption{Qualitative held-out Probe-B prediction on Pillow (target 002, \texttt{standard\_y}). Rows: mid frame, final frame, and per-pixel mean absolute RGB error at the final frame under one shared color scale, clipped at the 99.9th percentile so mid-range differences stay visible. Panels are cropped to the region that moves. Headers report target-specific sequence-level metrics over all 31 exported frames: foreground PSNR in dB / object-crop SSIM. CMA-ES uses predeclared seed 6401; aggregate direct-optimization results average all three seeds.}
\label{fig:pillow_qualitative}
\end{figure*}

Fig.~\ref{fig:pillow_qualitative} makes the same held-out-probe result visible in image space, showing a mid and a final exported frame of target 002 under unseen \texttt{standard\_y}; no frame or Probe-B signal is used for fitting. Local ridge reaches 41.20~dB foreground PSNR and 0.998 object-crop SSIM, compared with 32.93~dB/0.977 for global ridge, 31.75~dB/0.974 for response KNN, 26.54~dB/0.929 for the fixed CMA-ES seed shown, and 19.41~dB/0.892 for the default material. Because the deformation is small relative to the static basket, the renders are visually close; the shared-scale error maps localize the mismatch to the deforming seat cushion and reproduce the same method ordering as the quantitative table.

Across three predeclared alternative target splits, local ridge has lower mean parameter error than the strongest non-oracle method in all three. A stricter target-wise non-inferiority gate passes two of three splits: the failed split has only $3/5$ non-inferior targets, while the others have $4/5$ and $5/5$. All six split--protocol Probe-B response gates pass. We therefore claim a stable mean advantage, not universal target-wise dominance. Appendix Fig.~\ref{fig:offline_ablations} further evaluates candidate-library size and local-ridge $(k,\alpha)$ sensitivity using only the frozen response features, without additional MPM or rendering.

\subsection{Object-Specific Repeatability}
\label{sec:exp:objects}

Results on a single deformable asset may conflate estimator behavior with
asset-specific geometry and contact dynamics. We therefore repeat the complete
calibration-and-prediction protocol on two additional assets, keeping the
response descriptor and local-ridge configuration fixed while constructing an
independent response library for each asset. This experiment evaluates whether
the object-specific procedure remains effective across distinct assets; it is
not a cross-object transfer setting.

\begin{figure}[!t]
\centering
\includegraphics[width=0.94\linewidth]{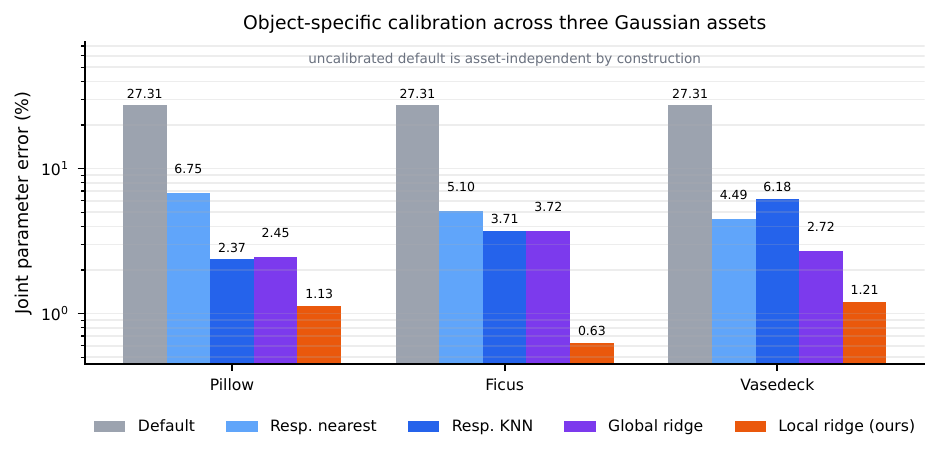}
\caption{Object-specific material calibration. The same estimator form is used, but each Gaussian asset has its own response library. Local ridge remains the best deployable method on Pillow, Ficus, and Vasedeck.}
\label{fig:objects}
\end{figure}

Fig.~\ref{fig:objects} repeats calibration independently for three geometrically distinct assets. Local ridge obtains 1.13\% error on Pillow, 0.63\% on Ficus, and 1.21\% on Vasedeck. On the Vasedeck \texttt{standard\_x}$\rightarrow$\texttt{standard\_z} test, its Probe-B trajectory RMSE is $1.37\times10^{-5}$, versus $3.11\times10^{-5}$ for the strongest non-oracle response-nearest result, and it is non-inferior on all five targets. Ficus yields 0.63\% scale error versus 3.71\% for KNN/global ridge and 27.31\% for the default material. Across the three evaluated assets, the same calibration procedure remains
effective when instantiated with an object-specific response library.

Rendered response follows the dynamics metrics. Across six pillow protocols, local ridge improves mean foreground PSNR from 55.50\,dB (global ridge) to 59.53\,dB and object-crop SSIM from 0.9968 to 0.9982. On Ficus, its mean foreground PSNR is 41.87\,dB versus 26.80\,dB for global ridge. All reported PSNR and SSIM values compare simulator-rendered predictions
with simulator-rendered targets and therefore do not measure agreement with
real video.

\begin{figure*}[!t]
\centering
\includegraphics[width=0.96\textwidth]{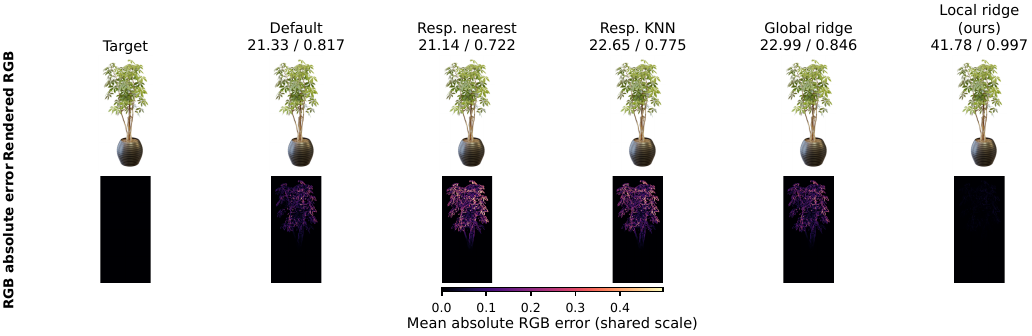}
\caption{Qualitative held-out Probe-B prediction on the median-error Ficus target ($s_E=1.05$, $s_\rho=1.10$, \texttt{standard\_y}). Rows show the final rendered frame and its mean absolute RGB error on one shared scale. Headers report target-specific foreground PSNR in dB / foreground SSIM averaged over all 30 non-initial exported frames. The target is selected only by the median local-ridge trajectory RMSE among all five held-out targets, before inspecting these visual scores. This is an independently calibrated object-specific loop, not cross-object transfer.}
\label{fig:ficus_qualitative}
\end{figure*}

Fig.~\ref{fig:ficus_qualitative} visualizes the objectively selected median Ficus case: we sort the five held-out targets by local-ridge Probe-B trajectory RMSE and show the middle one, $(s_E,s_\rho)=(1.05,1.10)$. This rule does not inspect PSNR/SSIM or baseline ordering. Averaged over exported frames, local ridge reaches 41.78\,dB/0.997 foreground PSNR/SSIM, versus 22.99\,dB/0.846 for global ridge, 22.65\,dB/0.775 for response KNN, 21.14\,dB/0.722 for response nearest, and 21.33\,dB/0.817 for the default material. The residual maps show that the remaining disagreement is concentrated on leaf boundaries and thin branches, where small trajectory errors cause large image-space differences.

\subsection{Diagnostic Boundaries}
\label{sec:exp:diagnostics}

\subsubsection{Observation Ladder}
\label{sec:exp:observation}

Everything above uses simulator-exported particle state, which no camera can provide. This diagnostic addresses the observation component of question (3): by removing one privilege at a time we locate \emph{which} property of the observation the method actually depends on, so that the distance to a deployable visual front end is a measured quantity rather than a guess.

\begin{figure}[!t]
\centering
\includegraphics[width=0.98\linewidth]{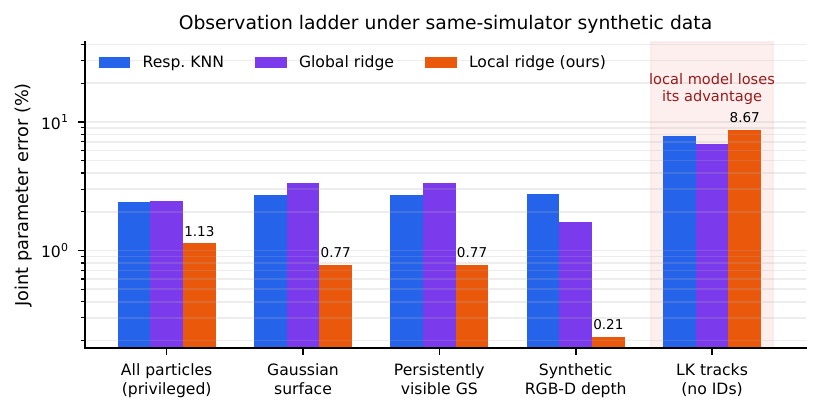}
\caption{Synthetic observation ladder. Clean surface and depth observations preserve useful calibration evidence under the same simulator, but the local estimator loses its advantage with ID-free LK tracks. The apparently lower clean-depth error is an empirical result for this split and feature construction, not a general claim that less information is intrinsically better.}
\label{fig:visual_ladder}
\end{figure}

The main estimator receives all exported particles, which is privileged state. To localize the gap to deployable vision, we progressively replace it with Gaussian surface particles, persistently visible Gaussians, clean synthetic RGB-D depth, and ID-free LK trajectories. Fig.~\ref{fig:visual_ladder} reports the result under the same simulator and target split. Local-ridge error is 1.13\% with all particles, 0.77\% with Gaussian surface or persistently visible Gaussians, and 0.21\% with clean synthetic depth. These values show that the selected descriptors can be recovered from clean surface observations in this controlled setting; they do not establish real RGB-D performance.

The conclusion changes at the final step. With LK tracks and no persistent particle identity, local-ridge error rises to 8.67\%, worse than global ridge (6.73\%) and KNN (7.76\%). The current bottleneck is therefore not simply whether internal fill particles are observed; robust correspondence and observation noise are unresolved. Synthetic-depth perturbations reinforce this boundary: local error grows from 0.21\% clean to 2.65\% at noise level 0.001, 16.25\% at 0.005, and 28.26\% at 0.01.

\subsubsection{Alternative Inverse Routes and External Diagnostics}
\label{sec:exp:alternatives}

The previous experiments compare estimators that share our library. This diagnostic tests the library route against two families that solve the inverse problem from different information---per-scene optimization against a video, and learned prediction from appearance or passive dynamics---and then explains, structurally, why the passive route loses information that a prescribed probe retains. Because their evidence and parameter semantics differ, these routes define boundaries rather than an additional same-information ranking.

\begin{figure*}[!t]
\centering
\includegraphics[width=0.9\textwidth]{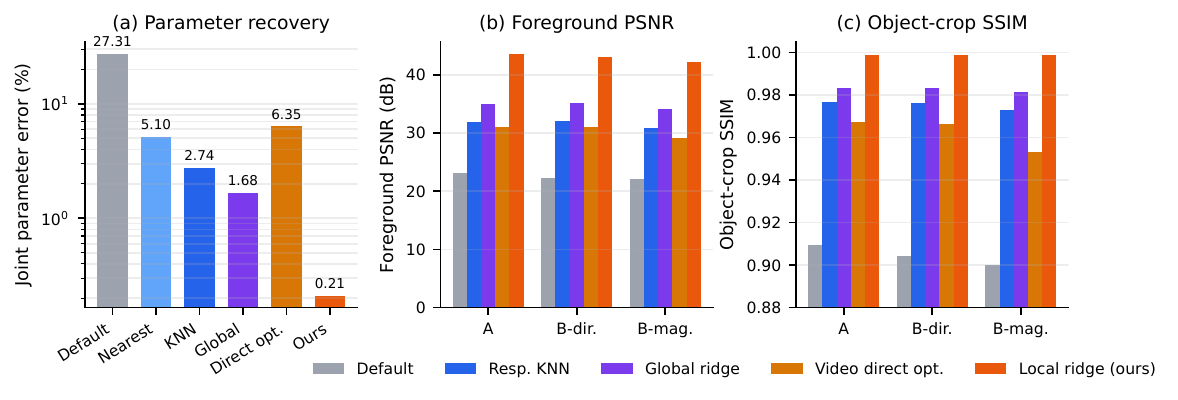}
\caption{Diagnostic comparison with fixed-budget per-scene video optimization. (a) Parameter recovery; (b) foreground PSNR; (c) object-crop SSIM. Both routes use simulator-generated RGB-D Probe-A evidence, and Probe B is excluded from fitting. The response-library estimator is more accurate under this budget, while requiring an offline object-specific library.}
\label{fig:directopt}
\end{figure*}

\paragraph{Direct optimization from one Probe-A video.}
To test whether the response library is necessary, we optimize $(s_E,s_\rho)$ directly against a single simulator-rendered RGB-D Probe-A video using CMA-ES. The optimizer receives no particle IDs and uses 30 forward calls per target with three predeclared seeds. Parameters are then frozen and evaluated on \texttt{standard\_y} and \texttt{strong\_x}. Fig.~\ref{fig:directopt} shows that fixed-budget direct optimization improves substantially over default parameters (6.35\% versus 27.31\% joint error), but remains behind response-library local calibration (0.21\% on the same synthetic-depth observation track). On Probe A, local calibration achieves 43.49\,dB PSNR versus 31.10\,dB for CMA-ES; the ordering persists on both Probe B protocols. This comparison has complementary computational costs: our method amortizes a 54-simulation object-specific library, whereas direct optimization incurs per-target simulator calls.

\paragraph{Visual-prior and passive-video systems.}
We additionally run PhysGM and ReconPhys as diagnostic external comparisons. Their native physical parameterizations, training distributions, and solvers differ from our MPM scales, so their outputs are adapted to the same initial asset and common downstream MPM rather than treated as same-information baselines. PhysGM observes four static views of the pillow scene and returns one material prediction for five visually identical targets; its common-adapter Probe-A response error is 11.42$\times$ ours and worse than the uncalibrated default on all five targets. ReconPhys observes passive video and is calibrated to our candidate range; it improves joint parameter error from the default's 34.99\% to 20.29\% and beats default response on $4/5$ targets, but its Probe-A and unseen-Probe-B video errors remain approximately 2.6$\times$ and 2.9$\times$ ours. Appendix Fig.~\ref{fig:external_qualitative} shows the corresponding common-adapter motion on a representative difficult target. These results support the value of known dynamic excitation in this controlled scene.

\paragraph{Why passive observation under-discriminates here.}
This discrepancy reflects an identifiability limitation of the passive observation model, rather than only a difference in estimator accuracy; it has a structural cause that we can state and then check. In the spring--mass formulation used by ReconPhys, the released implementation computes spring and damping forces that are linear in $k$ and $d$, adds gravity as $m\mathbf{g}$, and integrates $\mathbf{v}\!\leftarrow\!\mathbf{v}+\mathbf{F}\Delta t/m$; the released configuration resolves ground contact by a mass-independent kinematic projection. The rescaling $(m,k,d)\mapsto(\alpha m,\alpha k,\alpha d)$ therefore leaves every particle trajectory---and hence every rendered frame---unchanged. Under a passive drop, only ratios such as $k/m$ are identifiable, and absolute mass is not observable at all. This invariance arises from the observation setting and parameterization rather than from an implementation error. A mass-dependent penalty contact model, for example, could break this scaling invariance.

\begin{figure*}[!t]
\centering
\begin{minipage}[t]{0.47\textwidth}
\centering
\textbf{(a)}\\[-0.5ex]
\includegraphics[width=\linewidth]{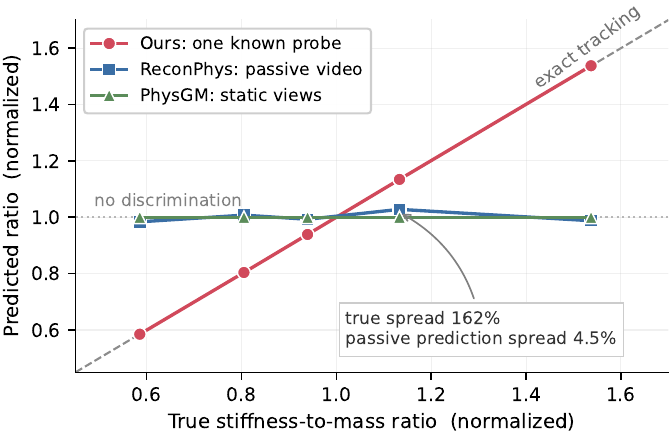}
\end{minipage}\hfill
\begin{minipage}[t]{0.47\textwidth}
\centering
\textbf{(b)}\\[-0.5ex]
\includegraphics[width=\linewidth]{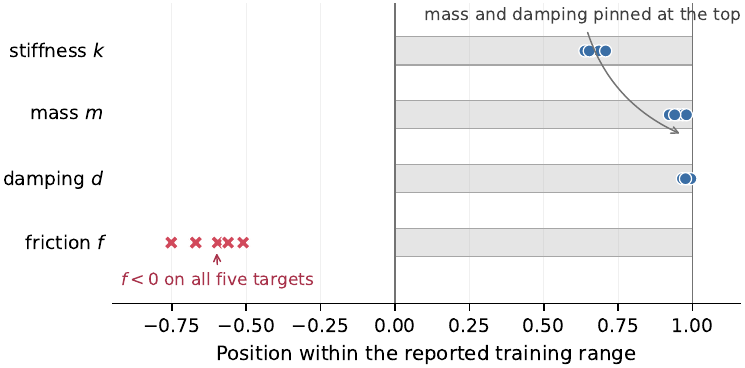}
\end{minipage}
\caption{Passive-observation diagnostics on five held-out targets. (a) Discriminative range after per-series mean normalization. (b) Released ReconPhys native outputs relative to its reported training ranges; crosses are inadmissible. These panels diagnose range and saturation under different parameterizations, not like-for-like parameter accuracy.}
\label{fig:teaser}
\label{fig:degeneracy}
\end{figure*}

Fig.~\ref{fig:teaser} checks the consequence on our five held-out targets. Panel (a) shows that the true ratio $s_E/s_\rho$ spans 162\% and our estimates span 163\%, whereas the released ReconPhys predictions span only 4.5\% in $k/m$ and PhysGM returns a single prediction because its four static views are identical across targets. Panel (b) adds the per-parameter view: the released outputs sit at the top of the reported training range for mass ($5.55$--$5.88$ within $[0.2,6.0]$) and damping ($4.83$--$4.97$ within $[0.1,5.0]$), and friction is negative on all five targets although the method defines $f\ge 0$. These observations are consistent with predictions dominated by the training prior rather than by target-specific evidence, and they are the concrete reason a known applied probe helps: the excitation is prescribed rather than inferred, so the scale direction that passive observation cannot resolve is fixed by construction.

\subsubsection{Discretization, Object, and Identifiability Boundaries}
\label{sec:exp:boundaries}

Sec.~\ref{sec:method:library} constructs each response library under a fixed
object-specific simulation contract. To characterize the estimator's
dependence on this contract, we vary one component at a time, evaluate the
resulting mismatch using the original library, and, where applicable, rebuild
a contract-matched library. These controlled comparisons distinguish
variations tolerated by the frozen library from those that require library
reconstruction.

\begin{figure*}[!t]
\centering
\includegraphics[width=0.9\textwidth]{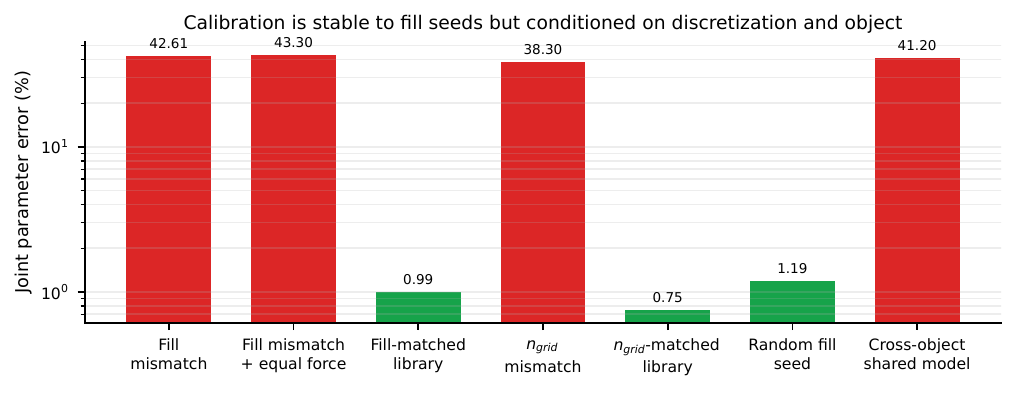}
\caption{Calibration boundaries for local ridge. Red bars violate the library contract; green bars preserve or rebuild it. Equalizing total force does not fix fill mismatch, while fill- and grid-matched libraries restore accuracy.}
\label{fig:boundaries}
\end{figure*}

Fig.~\ref{fig:boundaries} summarizes these controlled stress tests. Changing fill density while retaining the original library increases local-ridge error to 42.61\%; normalizing the total applied force does not resolve it (43.30\%). Rebuilding a fill-matched library restores 0.99\%. Changing MPM grid resolution gives 38.30\% error, while a grid-matched library restores 0.75\%. In contrast, random fill seeds at the same specification remain stable (1.19\% mean, 2.45\% maximum). Finally, leave-one-object-out use of a shared estimator fails with 38--45\% error. The method is therefore robust to stochastic fill realization but conditioned on the object and numerical discretization used to construct the library.

The five-dimensional descriptor is also not globally identifiable. Among 54 candidates (1,431 pairs), some distinct parameter pairs are close after descriptor compression, and local-neighborhood condition numbers have median 1,244 and maximum 6,539. Several compressed-space ambiguities become separable when the full trajectory is considered, showing that feature compression contributes additional ambiguity; this does not rule out physical non-identifiability under a fixed probe. 

%% file: sec/5._conclusion.tex
\section{Conclusion}
\label{sec:conclu}
We presented KnockGS, a controlled study of response-conditioned calibration for physics-integrated Gaussian assets. An object-specific library and a simple hard-neighborhood local ridge estimator recover two MPM material scales from a known Probe-A response; after write-back, the frozen estimate predicts held-out direction and magnitude probes more accurately than response retrieval, KNN, global ridge, and fixed default materials. The result repeats in object-specific loops on Pillow, Ficus, and Vasedeck, and is supported by Gaussian-associated-particle trajectories and rendered visual fidelity.

Equally important, the stress tests delimit the result. Clean synthetic surface and depth observations retain useful evidence, but ID-free visual tracks and depth perturbations degrade calibration. Fill or grid mismatch and cross-object sharing fail unless the response library is rebuilt under matching conditions.  Closing that gap requires stable visual correspondence, uncertainty-aware identification, broader physical states, and validation with measured force and object-level deformation.

\section{Discussion}
\label{sec:exp:scope}

The supported claim is controlled, object-specific, discretization-conditioned calibration of two MPM material scales from a known interaction, followed by unseen-interaction prediction in the same simulator family. The experiments do not establish real RGB/RGB-D input, real force feedback, measured material ground truth, sim-to-real transfer, a cross-object shared estimator, fill-unknown inversion, full multi-parameter material identification, damage prediction, or active-probe optimality. The observation ladder uses synthetic observations, the external routes use different native parameterizations, and the passive-identifiability result applies to the analyzed spring--mass/contact formulation; none should be read as a broader real-world or like-for-like accuracy claim.

The measured failures point to concrete future work. The degradation with ID-free tracks motivates robust visual correspondence, while sensitivity to synthetic-depth perturbations calls for uncertainty-aware estimation. Descriptor ambiguity under a fixed Probe A motivates active interaction selection and richer response representations, and the current two-scale state should be extended to richer physical parameterizations. Finally, closing the simulation-to-reality gap requires validation with measured interactions and real object deformation. These are future directions rather than capabilities of the current system.

%% file: sec/6._appendix.tex
\section{Response and Metric Details}
\label{app:metrics}

Let $\mathcal{I}_{\mathrm{GS}}$ denote Gaussian-associated particles with stable identity and $\mathcal{I}_{\mathrm{fill}}$ internal fill particles. The estimator's privileged input may use all exported particles, while the primary trajectory evaluation is restricted to $\mathcal{I}_{\mathrm{GS}}$. This avoids two confounds: internal fill particles are not directly renderable, and changing fill construction can change their count and correspondence. Exported states are recorded only at the configured output times, not at every internal MPM integration step.

For secondary curve metrics, normalized particle displacement is
\begin{equation}
\tilde d_i^t=\|\mathbf{x}_i^t-\mathbf{x}_i^0\|_2/D_{\mathrm{bbox}},
\end{equation}
and the RMS curve is $r^t=(|\mathcal{I}|^{-1}\sum_{i\in\mathcal{I}}(\tilde d_i^t)^2)^{1/2}$. We report
\begin{equation}
\mathrm{RMSE}_{\mathrm{curve}}=
\sqrt{\frac{1}{|\mathcal{T}_{\mathrm{out}}|}
\sum_t(\hat r^t-r^{\star,t})^2},
\end{equation}
plus $|\max_t\hat r^t-\max_t r^{\star,t}|$, the absolute difference of discrete AUC sums, and final-frame error. Curve metrics summarize temporal magnitude but do not replace the spatial trajectory metric.

Foreground PSNR is computed over the union of the target and predicted foreground masks, pooling RGB squared error and sample count over all 31 exported frames before conversion to dB. Object-crop SSIM is computed per frame within the bounding box of the same mask union, padded by five pixels, and then averaged across frames. Both target and prediction are rendered by the same Gaussian renderer and camera. They quantify same-simulator visual response fidelity and must not be interpreted as performance on captured real video.

\clearpage
\section{Probe Definitions and Separation}
\label{app:probes}

The candidate set comprises 54 predeclared pairs, not a Cartesian grid. Its coordinate levels are $s_E\in\{0.50,0.55,0.65,0.75,0.85,0.95,1.00,1.05,1.15\}$ and $s_\rho\in\{0.75,1.00,1.10,1.20,1.25,1.35,1.50,1.65,1.75\}$. The five targets are $(0.60,1.65)$, $(0.70,1.20)$, $(0.80,1.60)$, $(0.95,1.35)$, and $(1.05,1.10)$; none is one of the 54 candidate pairs. Material scales multiply the frozen object-specific base values in Table~\ref{tab:object_contract}. Candidate-only standardization and validation never use these targets.

\begin{table*}[!t]
\centering
\caption{Frozen object-specific simulation contract. Counts are physical-state particles used by the estimator; Pillow contains 624,324 Gaussian-associated particles and 71,660 internal fill particles. Primary Pillow trajectory evaluation samples at most 100,000 Gaussian-associated particles with seed 42. ``Exports'' includes the initial state.}
\label{tab:object_contract}
\small
\begin{tabular}{lcccll}
\toprule
Object & Base $E$ & Base $\rho$ & $\nu$ & Physical-state population & Exports / horizon \\
\midrule
Pillow & $10{,}000$ & $2{,}000$ & 0.30 & 695,984 (GS + fixed-seed fill) & 31 / 0.60\,s \\
Ficus & $2{\times}10^6$ & 200 & 0.40 & 171,553 Gaussian-associated & 31 / 1.20\,s \\
Vasedeck & $10{,}000$ & 40 & 0.30 & 55,364 Gaussian-associated & 31 / 0.60\,s \\
\bottomrule
\end{tabular}
\end{table*}

All three objects use an internal MPM substep of $10^{-4}$\,s. Pillow uses $n_{\mathrm{grid}}=100$ and a $128^3$ fill grid with at most four particles per cell and frozen fill seed 60042. Ficus scales both its global material and its configured branch/leaf material region by the same $(s_E,s_\rho)$ pair. Vasedeck uses $n_{\mathrm{grid}}=120$. Geometry, coordinate transforms, support constraints, gravity, damping, friction, camera, and renderer remain fixed within each object-specific library.

\begin{table*}[!t]
\centering
\caption{Exact probe values. $\mathbf f$ is the force vector per selected particle; contact boxes are reported as center $\mathbf c$ / half-extent $\mathbf h$. One MPM step is 0.1\,ms. Global Pillow and Ficus probes select the full physical-state population. The duration row preserves the nominal force--time integral of \texttt{standard\_x}; the two local-contact rows preserve total force and impulse relative to each other.}
\label{tab:probe_values}
\footnotesize
\setlength{\tabcolsep}{3pt}
\resizebox{\textwidth}{!}{%
\begin{tabular}{lllclrr}
\toprule
Protocol & Object / role & $\mathbf f$ & Contact $\mathbf c/\mathbf h$ & Steps / duration & Selected $N$ & Exports \\
\midrule
\texttt{standard\_x} & Pillow A & $(-0.18,0,0)$ & global & 1 / 0.1\,ms & 695,984 & 31 at 20\,ms \\
\texttt{standard\_y} & Pillow B-direction & $(0,-0.18,0)$ & global & 1 / 0.1\,ms & 695,984 & 31 at 20\,ms \\
\texttt{strong\_x} & Pillow B-magnitude & $(-0.36,0,0)$ & global & 1 / 0.1\,ms & 695,984 & 31 at 20\,ms \\
\texttt{duration\_x\_fine} & Pillow B-duration & $(-0.0009,0,0)$ & global & 200 / 20\,ms & 695,984 & 121 at 5\,ms \\
\texttt{local\_center\_x} & Pillow contact ref. & $(-0.18,0,0)$ & $(1,1,1.28)/(0.16,0.16,0.16)$ & 1 / 0.1\,ms & 94,657 & 31 at 20\,ms \\
\texttt{shifted\_contact\_x} & Pillow B-location & $(-0.226245,0,0)$ & $(1,1.12,1.28)/(0.16,0.16,0.16)$ & 1 / 0.1\,ms & 75,309 & 31 at 20\,ms \\
\texttt{standard\_x/y} & Ficus A / B-direction & $(-0.18,0,0)/(0,-0.18,0)$ & global & 1 / 0.1\,ms & 171,553 & 31 at 40\,ms \\
\texttt{standard\_x/z} & Vasedeck A / B-direction & $(-0.09,0,0)/(0,0,-0.09)$ & $(1.05,0.9,0.92)/(0.2,0.14,0.2)$ & 1 / 0.1\,ms & 4,464 & 31 at 20\,ms \\
\bottomrule
\end{tabular}
}
\end{table*}

All parameters are estimated using only \texttt{standard\_x}. Probe-B responses are generated only after the estimate, descriptor definition, neighborhood size, and regularization have been frozen. The candidate library contains Probe-B rollouts for efficient evaluation, but no Probe-B feature participates in the estimator.

\section{Target-wise Results and Variability}
\label{app:targetwise}

Table~\ref{tab:v59_mean_sd} reports sample standard deviation across the five held-out targets. This quantifies target-to-target variation, not simulator noise: fixed default, retrieval, KNN, and both ridge estimators are deterministic once the library, fill seed, and evaluation sample are frozen. We do not treat 31 frames or up to 100,000 particles as independent replicates.

\begin{table*}[!t]
\centering
\caption{Pillow mean $\pm$ sample standard deviation across five held-out targets. Trajectory columns are in units of $10^{-5}$; lower is better.}
\label{tab:v59_mean_sd}
\footnotesize
\setlength{\tabcolsep}{5pt}
\begin{tabular}{lrrrr}
\toprule
Method & Joint error (\%) & Probe A & B-direction & B-magnitude \\
\midrule
Fixed default & $27.313\pm17.577$ & $368.195\pm248.079$ & $350.172\pm240.806$ & $440.730\pm273.547$ \\
Response nearest & $6.746\pm2.550$ & $37.219\pm11.919$ & $35.328\pm11.008$ & $51.876\pm14.660$ \\
Response KNN & $2.374\pm1.065$ & $14.491\pm8.842$ & $14.119\pm8.791$ & $20.203\pm9.086$ \\
Global ridge & $2.449\pm1.594$ & $17.476\pm11.528$ & $16.874\pm10.905$ & $24.660\pm16.290$ \\
\rowcolor{oursgray}\textbf{Local ridge (ours)} & $\mathbf{1.130\pm0.605}$ & $\mathbf{4.498\pm1.965}$ & $\mathbf{4.424\pm1.892}$ & $\mathbf{7.786\pm3.858}$ \\
Parameter-nearest oracle & $4.447\pm0.628$ & $62.827\pm24.337$ & $59.837\pm24.313$ & $71.438\pm24.884$ \\
\bottomrule
\end{tabular}
\end{table*}

\begin{table*}[!t]
\centering
\caption{All Pillow held-out targets and methods. $e_\theta$ is joint parameter error in percent; trajectory columns are in units of $10^{-5}$. No target or method outcome is omitted.}
\label{tab:v59_targetwise}
\scriptsize
\setlength{\tabcolsep}{4pt}
\renewcommand{\arraystretch}{0.80}
\begin{tabular}{llrrrr}
\toprule
Target $(s_E,s_\rho)$ & Method & $e_\theta$ & Probe A & B-direction & B-magnitude \\
\midrule
$(0.60,1.65)$ & Fixed default & 53.030 & 715.049 & 690.177 & 814.549 \\
 & Resp. nearest & 7.197 & 28.205 & 27.441 & 40.864 \\
 & Resp. KNN & 1.417 & 28.593 & 28.326 & 29.045 \\
 & Global ridge & 1.496 & 28.680 & 28.274 & 29.257 \\
 & \textbf{Local ridge} & \textbf{0.224} & \textbf{1.618} & \textbf{1.574} & \textbf{1.846} \\
 & Param.-nearest oracle & 4.167 & 91.944 & 90.996 & 93.126 \\
\midrule
$(0.70,1.20)$ & Fixed default & 29.762 & 349.973 & 325.707 & 401.730 \\
 & Resp. nearest & 3.571 & 56.453 & 53.032 & 62.402 \\
 & Resp. KNN & 2.113 & 9.326 & 9.476 & 16.547 \\
 & Global ridge & \textbf{0.657} & 6.695 & 6.233 & \textbf{7.580} \\
 & Local ridge & 1.371 & \textbf{5.592} & \textbf{5.647} & 10.026 \\
 & Param.-nearest oracle & 3.571 & 56.453 & 53.032 & 62.402 \\
\midrule
$(0.80,1.60)$ & Fixed default & 31.250 & 475.247 & 454.395 & 577.379 \\
 & Resp. nearest & 4.687 & 26.407 & 24.899 & 32.067 \\
 & Resp. KNN & 1.812 & 6.405 & 6.076 & 10.026 \\
 & Global ridge & 3.354 & 11.735 & 12.082 & 21.260 \\
 & \textbf{Local ridge} & \textbf{0.993} & \textbf{3.457} & \textbf{3.576} & \textbf{6.191} \\
 & Param.-nearest oracle & 4.687 & 26.407 & 24.899 & 32.067 \\
\midrule
$(0.95,1.35)$ & Fixed default & 15.595 & 248.238 & 231.909 & 323.815 \\
 & Resp. nearest & 8.967 & 37.628 & 35.814 & 58.202 \\
 & Resp. KNN & 2.353 & 10.792 & 10.203 & 14.869 \\
 & Global ridge & 2.039 & 9.118 & 8.857 & 15.126 \\
 & \textbf{Local ridge} & \textbf{1.881} & \textbf{6.573} & \textbf{6.340} & \textbf{11.588} \\
 & Param.-nearest oracle & 5.263 & 75.183 & 71.076 & 81.811 \\
\midrule
$(1.05,1.10)$ & Fixed default & 6.926 & 52.467 & 48.675 & 86.176 \\
 & Resp. nearest & 9.307 & 37.403 & 35.456 & 65.846 \\
 & Resp. KNN & 4.173 & 17.339 & 16.512 & 30.530 \\
 & Global ridge & 4.697 & 31.152 & 28.924 & 50.079 \\
 & \textbf{Local ridge} & \textbf{1.178} & \textbf{5.249} & \textbf{4.982} & \textbf{9.279} \\
 & Param.-nearest oracle & 4.545 & 64.145 & 59.183 & 87.783 \\
\bottomrule
\end{tabular}
\renewcommand{\arraystretch}{1.0}
\end{table*}

The per-target table exposes the only primary-protocol exception: on $(0.70,1.20)$, global ridge has lower parameter and B-magnitude errors, while local ridge remains lower on Probe A and B-direction. This is why the paper claims the best mean and reports target-wise gates rather than universal dominance.

\FloatBarrier
\section{Duration and Contact-Shift Diagnostics}
\label{app:auxiliary_probes}

The fine duration experiment reuses the five frozen \texttt{standard\_x} estimates and changes only the force time profile. It resolves the 20\,ms actuation with 5\,ms exports (121 states over 0.6\,s), rather than the 20\,ms exports used by the main benchmark. The 0.0009-per-particle force is applied for 200 internal steps, giving the same nominal force--time integral as the 0.18 force applied for one step. The two target protocols are measurably distinct (mean target-to-target protocol trajectory RMSE $8.998\times10^{-5}$).

\begin{center}
\centering
\captionof{table}{Five-target duration-shift result, mean $\pm$ sample standard deviation; values are $\times10^{-5}$. Parameters are estimated only from \texttt{standard\_x}.}
\label{tab:duration_results}
\small
\begin{tabular}{lrr}
\toprule
Method & Trajectory RMSE & Curve RMSE \\
\midrule
Response KNN & $8.224\pm5.127$ & $7.886\pm9.484$ \\
Global ridge & $9.895\pm6.541$ & $9.882\pm9.230$ \\
\rowcolor{oursgray}\textbf{Local ridge} & $\mathbf{2.503\pm1.074}$ & $\mathbf{0.978\pm0.270}$ \\
\bottomrule
\end{tabular}
\end{center}

For contact location, the existing completed experiment is deliberately reported as a single-target diagnostic, not as five-target evidence. On the predeclared $(0.95,1.35)$ target, \texttt{local\_center\_x} and \texttt{shifted\_contact\_x} use equal total force and impulse; the latter shifts the contact-box center by 0.12 along $y$ and raises per-particle force to compensate for the smaller selected set. This is a location-only comparison against the local-center reference, not against global \texttt{standard\_x}.

\begin{center}
\centering
\captionof{table}{Single-target contact-location diagnostic; values are $\times10^{-5}$. The same Probe-A parameter estimate is frozen for both contacts.}
\label{tab:contact_results}
\small
\begin{tabular}{llrr}
\toprule
Contact & Method & Trajectory RMSE & Curve RMSE \\
\midrule
Local center & Global ridge & 4.115 & 3.254 \\
Local center & \textbf{Local ridge} & \textbf{0.281} & \textbf{0.221} \\
Shifted & Global ridge & 4.115 & 3.255 \\
Shifted & \textbf{Local ridge} & \textbf{0.281} & \textbf{0.221} \\
\bottomrule
\end{tabular}
\end{center}

Local ridge wins on all five duration targets and on both contact locations for the diagnostic target. The duration result supports a five-target protocol-shift claim; the contact result supports only a controlled representative demonstration and is retained with that limitation to avoid selective overstatement.

\section{Multiple Splits and Numerical Tests}
\label{app:robustness}

\begin{table}[!t]
\centering
\caption{Three independent five-target splits. The strict gate requires a lower mean than the strongest non-oracle method and at least four of five target-wise non-inferior results.}
\label{tab:multisplit}
\small
\begin{tabular}{lrrrr}
\toprule
Split & Local (\%) & Strongest non-oracle (\%) & Non-inferior & Gate \\
\midrule
1 & 1.39 & 2.60 & 3/5 & fail \\
2 & 1.28 & 3.47 & 4/5 & pass \\
3 & 0.76 & 4.30 & 5/5 & pass \\
\bottomrule
\end{tabular}
\end{table}

Local ridge has the lower mean on every split, while the strict target-wise gate passes only two. Probe-B resimulation passes all six split--protocol gates. We retain both facts to distinguish average performance from universal dominance.

The fill and grid experiments separate random realization from a changed numerical model. Random fill seeds preserve the same fill specification and give 1.19\% mean error. Changing fill density or $n_{\mathrm{grid}}$ changes the response distribution and produces 42.61\% and 38.30\% error with the original library. Rebuilding matched libraries restores 0.99\% and 0.75\%, respectively. Equal-total-force normalization alone leaves fill mismatch at 43.30\%, so the failure cannot be attributed only to the number of selected particles or total impulse.

\section{Offline Estimator Sensitivity}
\label{app:offline_sensitivity}

We test two estimator design choices using only the frozen response-feature tables; no MPM rollout, rendering, or new response generation is performed. The audit covers 2,375 offline predictions: 605 library-size trials, 150 held-out $(k,\alpha)$ trials, and 1,620 candidate-only leave-one-out trials. Every result is finite, no estimator fallback is triggered, and the 54 candidates have zero parameter-pair overlap with the five held-out targets.

\begin{figure*}[!t]
\centering
\includegraphics[width=0.98\textwidth]{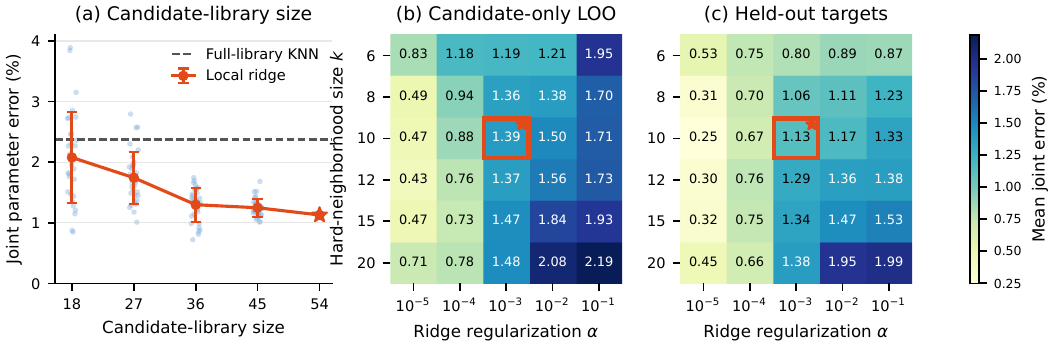}
\caption{Offline estimator sensitivity; lower is better. (a) Candidate-library size with the local-ridge configuration frozen at $k=10$, $\alpha=10^{-3}$. For 18/27/36/45 candidates, dots are the five-target mean from each of 30 predeclared nested random subsets and error bars show standard deviation across subset means; the full 54-candidate library is unique. The dashed line is the full-library response-KNN error. (b) Candidate-only leave-one-out sensitivity over all 54 candidates. (c) Sensitivity on the five held-out targets, used only as a post-hoc diagnostic. The red box and star mark the frozen paper configuration; neither heatmap is used to retune the reported main result.}
\label{fig:offline_ablations}
\end{figure*}

Increasing the library from 18 to 54 candidates reduces mean joint error from 2.078\% to 1.130\%; the standard deviation caused by subset selection contracts from 0.757\% at 18 candidates to 0.151\% at 45 candidates and vanishes for the unique full library. Thus the larger library improves both accuracy and coverage stability. Even the 18-candidate mean remains below the 2.374\% full-library response-KNN baseline, indicating that the local model's advantage is not confined to the densest library.

The frozen $k=10$, $\alpha=10^{-3}$ configuration exactly reproduces the reported 1.130\% held-out mean. Weak regularization is consistently more accurate in these nearly noise-free synthetic features: candidate-only leave-one-out is lowest at $k=12$, $\alpha=10^{-5}$ (0.426\%), while the held-out diagnostic is lowest at $k=10$, $\alpha=10^{-5}$ (0.249\%). We retain the predeclared main configuration rather than retrospectively replacing it. Across all 30 held-out grid cells, mean error remains below the 2.374\% response-KNN result, supporting a method-family advantage while also revealing sensitivity to regularization strength.

\section{Observation Ladder Details}
\label{app:observation}

The observation ladder holds the asset, target split, simulator, and Probe A fixed. Its five stages are:
\begin{enumerate}
  \item all simulator particles (privileged state);
  \item Gaussian-associated surface particles;
  \item Gaussian-associated particles that remain visible;
  \item clean synthetic RGB-D depth projected into 3D;
  \item LK optical-flow tracks without persistent simulator IDs.
\end{enumerate}
The first four maintain exact or clean geometric correspondence supplied by the synthetic pipeline. The last is closer to a practical image tracker and breaks that correspondence. Consequently, the ladder identifies a likely bottleneck but is not a substitute for a captured RGB-D benchmark.

\section{Alternative-Route Diagnostics}
\label{app:external}

\begin{table*}[!t]
\centering
\caption{Diagnostic comparisons with methods using different information or parameter semantics. These rows are not a same-information leaderboard. ``Ratio to ours'' uses the corresponding common-adapter response/video metric.}
\label{tab:external}
\footnotesize
\setlength{\tabcolsep}{5pt}
\begin{tabular}{p{2.5cm}p{3.4cm}p{2.1cm}p{2.1cm}p{4.4cm}}
\toprule
Route & Input & Joint parameter error & Response ratio to ours & Interpretation boundary \\
\midrule
PhysGM & four static views & 5279.88\% & 11.42$\times$ (Probe A) & predicted Fabric/high stiffness; scene and parameter semantics may be out of distribution \\
ReconPhys adapter & passive video & 20.29\% & 2.6$\times$ A; 2.9$\times$ B & improves over default on 4/5 targets; native solver/parameterization differs \\
CMA-ES direct opt. & synthetic RGB-D Probe A & 6.35\% & PSNR 31.10 vs. 43.49\,dB & 30 calls per target, three seeds; fixed-budget, not exhaustive optimization \\
Local ridge (ours) & synthetic depth descriptor + library & 0.21\% & 1.0$\times$ & requires a 54-candidate offline library for the same object/discretization \\
\bottomrule
\end{tabular}
\end{table*}

PhysGM outputs a single parameter set for the five visually identical pillow targets; ReconPhys is calibrated into the available candidate range before common-adapter evaluation. These adaptations are useful for testing the proposition that known dynamic excitation carries information missing from appearance or passive input, but they do not establish across-paper superiority. The CMA-ES experiment is closer to the current target protocol because it directly fits Probe-A synthetic video and freezes parameters before Probe B; it therefore serves as the main alternative inverse baseline.

\begin{table*}[!t]
\centering
\caption{Target-wise CMA-ES direct-video optimization, mean $\pm$ sample standard deviation over predeclared seeds 6401, 6402, and 6403. Joint error is in percent and trajectory errors are $\times10^{-3}$. Each seed uses 30 Probe-A simulator calls; Probe B is evaluation-only.}
\label{tab:cmaes_targetwise}
\footnotesize
\begin{tabular}{lrrrr}
\toprule
Target $(s_E,s_\rho)$ & Joint error & Probe A & B-direction & B-magnitude \\
\midrule
$(0.95,1.35)$ & $7.58\pm6.80$ & $0.371\pm0.222$ & $0.357\pm0.218$ & $0.580\pm0.437$ \\
$(0.60,1.65)$ & $8.06\pm6.40$ & $0.420\pm0.310$ & $0.433\pm0.322$ & $0.637\pm0.471$ \\
$(1.05,1.10)$ & $5.10\pm3.74$ & $0.293\pm0.173$ & $0.277\pm0.165$ & $0.411\pm0.225$ \\
$(0.80,1.60)$ & $8.49\pm5.62$ & $0.289\pm0.198$ & $0.291\pm0.200$ & $0.532\pm0.365$ \\
$(0.70,1.20)$ & $2.50\pm1.93$ & $0.165\pm0.150$ & $0.157\pm0.136$ & $0.223\pm0.170$ \\
\bottomrule
\end{tabular}
\end{table*}

\begin{figure*}[!t]
\centering
\includegraphics[width=0.96\textwidth]{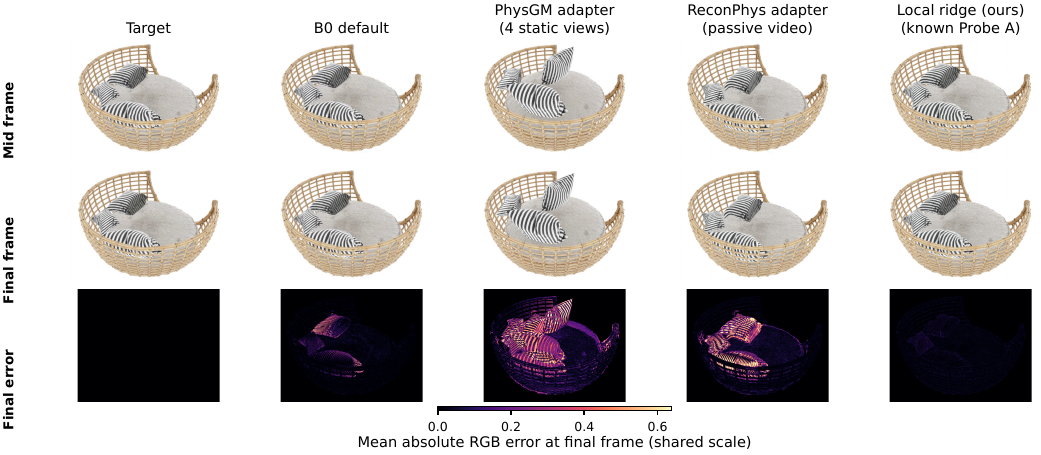}
\caption{External-route qualitative diagnostic on the representative held-out target $(s_E=1.05,s_\rho=1.1)$ under unseen direction Probe B1. Columns share the same initial asset, downstream MPM implementation, probe, camera, and exported times, but differ in the evidence used to assign material parameters. We show synchronized mid/final frames and final-frame absolute RGB error with one shared scale. PhysGM and ReconPhys remain diagnostic adapters with different native parameter semantics and training distributions; this is not a same-information leaderboard.}
\label{fig:external_qualitative}
\end{figure*}

Fig.~\ref{fig:external_qualitative} supplies the visual evidence corresponding to the PhysGM and ReconPhys rows of Table~\ref{tab:external}; the CMA-ES route is visualized separately in Fig.~\ref{fig:pillow_qualitative}. The displayed held-out target is selected as a representative difficult case, while all aggregate claims continue to use all five targets.

\section{Reproducibility and Claim Boundaries}
\label{app:reproducibility}

The released experiment contracts record the asset, 54 candidates, held-out targets, force definitions, contact regions, export times, particle subset, data split, feature standardization, neighborhood size, regularization, simulator configuration, and metric implementation. Result packages contain per-case CSV/JSON outputs and completion, leakage, missing-frame, NaN, and duplicate audits. No result is interpreted as real-material accuracy because target parameters and responses are generated by the same simulator.

The current evidence supports: (i) same-information local-calibration gain; (ii) object-specific Probe-A-to-B transfer; (iii) same-simulator visual-response fidelity; and (iv) an explicit characterization of observation and numerical-conditioning failures. It does not support real RGB/RGB-D identification, real force-feedback closure, sim-to-real transfer, a cross-object shared estimator, unknown-fill inversion, or a complete physical representation.

%% file: main.bib
@inproceedings{mildenhallnerf,
  title={{Nerf}: Representing scenes as neural radiance fields for view synthesis},
  author={Mildenhall, Ben and Srinivasan, Pratul P and Tancik, Matthew and Barron, Jonathan T and Ramamoorthi, Ravi and Ng, Ren},
  booktitle={European conference on computer vision},
  pages={405--421},
  year={2020},
  organization={Springer}
}

@article{kerbl3dgs,
  author  = {Kerbl, Bernhard and Kopanas, Georgios and Leimk{\"u}hler, Thomas and Drettakis, George},
  title   = {{3D} {G}aussian Splatting for Real-Time Radiance Field Rendering},
  journal = {ACM Transactions on Graphics},
  volume  = {42},
  number  = {4},
  year    = {2023}
}

@inproceedings{xiephysgaussian,
  author    = {Xie, Tianyi and Zong, Zeshun and Qiu, Yuxing and Li, Xuan and Feng, Yutao and Yang, Yin and Jiang, Chenfanfu},
  title     = {{PhysGaussian}: Physics-Integrated {3D} {G}aussians for Generative Dynamics},
  booktitle = {Proc. IEEE/CVF Conference on Computer Vision and Pattern Recognition (CVPR)},
  year      = {2024}
}

@article{sulskympm,
  author  = {Sulsky, Deborah and Chen, Zhen and Schreyer, Howard L.},
  title   = {A Particle Method for History-Dependent Materials},
  journal = {Computer Methods in Applied Mechanics and Engineering},
  volume  = {118},
  number  = {1--2},
  pages   = {179--196},
  year    = {1994}
}

@article{jiangmpm,
  author  = {Jiang, Chenfanfu and Schroeder, Craig and Selle, Andrew and Teran, Joseph and Stomakhin, Alexey},
  title   = {The Affine Particle-in-Cell Method},
  journal = {ACM Transactions on Graphics},
  volume  = {34},
  number  = {4},
  year    = {2015}
}

@inproceedings{linomniphysgs,
  author    = {Lin, Yuchen and Lin, Chenguo and Xu, Jianjin and Mu, Yadong},
  title     = {{OmniPhysGS}: {3D} Constitutive {G}aussians for General Physics-Based Dynamics Generation},
  booktitle = {Proc. International Conference on Learning Representations (ICLR)},
  year      = {2025}
}

@article{iphysgaussian,
  title={i-PhysGaussian: Implicit Physical Simulation for 3D Gaussian Splatting},
  author={Cao, Yicheng and Huang, Zhuo and Yao, Yu and Ying, Yiming and Dong, Daoyi and Liu, Tongliang},
  journal={arXiv preprint arXiv:2602.17117},
  year={2026}
}

@article{fastphysgs,
  title={FastPhysGS: Accelerating Physics-based Dynamic 3DGS Simulation via Interior Completion and Adaptive Optimization},
  author={Ma, Yikun and Li, Yiqing and Ye, Jingwen and Wu, Zhongkai and Zhang, Weidong and Gao, Lin and Jin, Zhi},
  journal={arXiv preprint arXiv:2602.01723},
  year={2026}
}

@article{gaussianfluent,
  title={Gaussianfluent: Gaussian simulation for dynamic scenes with mixed materials},
  author={Huang, Bei and Chen, Yixin and Lu, Ruijie and Zeng, Gang and Zha, Hongbin and Pei, Yuru and Huang, Siyuan},
  journal={arXiv preprint arXiv:2601.09265},
  year={2026}
}

@inproceedings{physsplat,
  title={PhysSplat: Efficient physics simulation for 3D scenes via MLLM-guided Gaussian splatting},
  author={Zhao, Haoyu and Wang, Hao and Zhao, Xingyue and Fei, Hao and Wang, Hongqiu and Long, Chengjiang and Zou, Hua},
  booktitle={2025 IEEE/CVF International Conference on Computer Vision (ICCV)},
  pages={5242--5252},
  year={2025},
  organization={IEEE}
}

@inproceedings{gaussianproperty,
  title={Gaussianproperty: Integrating physical properties to 3d gaussians with lmms},
  author={Xu, Xinli and Ge, Wenhang and Qiu, Dicong and Chen, ZhiFei and Yan, Dongyu and Liu, Zhuoyun and Zhao, Haoyu and Zhao, Hanfeng and Zhang, Shunsi and Liang, Junwei and others},
  booktitle={2025 IEEE/CVF International Conference on Computer Vision (ICCV)},
  pages={7231--7240},
  year={2025},
  organization={IEEE}
}

@inproceedings{pugs,
  title={Pugs: Zero-shot physical understanding with gaussian splatting},
  author={Shuai, Yinghao and Yu, Ran and Chen, Yuantao and Jiang, Zijian and Song, Xiaowei and Wang, Nan and Zheng, Jv and Ma, Jianzhu and Yang, Meng and Wang, Zhicheng and others},
  booktitle={2025 IEEE International Conference on Robotics and Automation (ICRA)},
  pages={4478--4485},
  year={2025},
  organization={IEEE}
}

@inproceedings{physgs,
  title={Physgs: Bayesian-inferred gaussian splatting for physical property estimation},
  author={Chopra, Samarth and Liang, Jing and Seneviratne, Gershom and Manocha, Dinesh},
  booktitle={Proceedings of the IEEE/CVF Conference on Computer Vision and Pattern Recognition},
  pages={18980--18990},
  year={2026}
}

@inproceedings{physgm,
  title={Physgm: Large physical gaussian model for feed-forward 4d synthesis},
  author={Lv, Chunji and Chen, Zequn and Di, Donglin and Zhang, Weinan and Li, Hao and Wei, Chen and Lei, Yinjie and Li, Changsheng},
  booktitle={Proceedings of the IEEE/CVF Conference on Computer Vision and Pattern Recognition},
  pages={29855--29865},
  year={2026}
}

@inproceedings{pacnerf,
  author    = {Li, Xuan and Qiao, Yi-Ling and Chen, Peter Yichen and Jatavallabhula, Krishna Murthy and Lin, Ming and Jiang, Chenfanfu and Gan, Chuang},
  title     = {{PAC-NeRF}: Physics Augmented Continuum Neural Radiance Fields for Geometry-Agnostic System Identification},
  booktitle = {Proc. International Conference on Learning Representations (ICLR)},
  year      = {2023}
}

@inproceedings{gic,
  author    = {Cai, Junhao and Yang, Yuji and Yuan, Weihao and He, Yisheng and Dong, Zilong and Bo, Liefeng and Cheng, Hui and Chen, Qifeng},
  title     = {{G}aussian-Informed Continuum for Physical Property Identification and Simulation},
  booktitle = {Advances in Neural Information Processing Systems (NeurIPS)},
  year      = {2024}
}

@inproceedings{springgaus,
  author    = {Zhong, Licheng and Yu, Hong-Xing and Wu, Jiajun and Li, Yunzhu},
  title     = {Reconstruction and Simulation of Elastic Objects with Spring-Mass {3D} {G}aussians},
  booktitle = {Proc. European Conference on Computer Vision (ECCV)},
  year      = {2024}
}

@inproceedings{phystwin,
  title={Phystwin: Physics-informed reconstruction and simulation of deformable objects from videos},
  author={Jiang, Hanxiao and Hsu, Hao-Yu and Zhang, Kaifeng and Yu, Hsin-Ni and Wang, Shenlong and Li, Yunzhu},
  booktitle={2025 IEEE/CVF International Conference on Computer Vision (ICCV)},
  pages={7219--7230},
  year={2025},
  organization={IEEE}
}

@misc{reconphys,
  author = {Wang, Boyuan and Wang, Xiaofeng and Li, Yongkang and Zhu, Zheng and Chang, Yifan and Ye, Angen and Zhao, Guosheng and Ni, Chaojun and Huang, Guan and Ren, Yijie and Duan, Yueqi and Wang, Xingang},
  title  = {{ReconPhys}: Reconstruct Appearance and Physical Attributes from Single Video},
  note   = {arXiv:2604.07882},
  year   = {2026}
}

@article{projo4d,
  title={Projo4d: Progressive joint optimization for sparse-view inverse physics estimation},
  author={Rho, Daniel and Choi, Jun Myeong and Dey, Biswadip and Sengupta, Roni},
  journal={arXiv preprint arXiv:2506.05317},
  year={2025}
}

@inproceedings{luitendynamic3dgs,
  author    = {Luiten, Jonathon and Kopanas, Georgios and Leibe, Bastian and Ramanan, Deva},
  title     = {Dynamic {3D} {G}aussians: Tracking by Persistent Dynamic View Synthesis},
  booktitle = {Proc. International Conference on 3D Vision (3DV)},
  year      = {2024}
}

@inproceedings{ngff,
  title={Learning physics-grounded 4D dynamics with neural Gaussian force fields},
  author={Li, Shiqian and Shen, Ruihong and Ni, Junfeng and Pan, Chang and Zhang, Chi and Zhu, Yixin},
  booktitle={International Conference on Learning Representations},
  volume={2026},
  pages={81165--81207},
  year={2026}
}

@inproceedings{diffwind,
  title={DiffWind: Physics-Informed Differentiable Modeling of Wind-Driven Object Dynamics},
  author={Lei, Yuanhang and Zhao, Boming and Yang, Zesong and Li, Xingxuan and Cheng, Tao and Peng, Haocheng and Zhang, Ru and Huang, Siyuan and Shen, Yujun and Hu, Ruizhen and others},
  booktitle={International Conference on Learning Representations},
  volume={2026},
  pages={23469--23494},
  year={2026}
}

@inproceedings{manigaussian,
  author    = {Lu, Guanxing and Zhang, Shiyi and Wang, Ziwei and Liu, Changliu and Lu, Jiwen and Tang, Yansong},
  title     = {{ManiGaussian}: Dynamic {G}aussian Splatting for Multi-Task Robotic Manipulation},
  booktitle = {Proc. European Conference on Computer Vision (ECCV)},
  year      = {2024}
}

@inproceedings{splatsim,
  title={Splatsim: Zero-shot sim2real transfer of rgb manipulation policies using gaussian splatting},
  author={Qureshi, M Nomaan and Garg, Sparsh and Yandun, Francisco and Held, David and Kantor, George and Silwal, Abhisesh},
  booktitle={2025 IEEE International Conference on Robotics and Automation (ICRA)},
  pages={6502--6509},
  year={2025},
  organization={IEEE}
}

@inproceedings{adaptigraph,
  author    = {Zhang, Kaifeng and Li, Baoyu and Hauser, Kris and Li, Yunzhu},
  title     = {{AdaptiGraph}: Material-Adaptive Graph-Based Neural Dynamics for Robotic Manipulation},
  booktitle = {Proc. Robotics: Science and Systems (RSS)},
  year      = {2024}
}

@article{instantngp,
  title={Instant neural graphics primitives with a multiresolution hash encoding},
  author={M{\"u}ller, Thomas and Evans, Alex and Schied, Christoph and Keller, Alexander},
  journal={ACM Transactions on Graphics},
  volume={41},
  number={4},
  pages={1--15},
  year={2022}
}

@inproceedings{mipnerf360,
  title={Mip-{NeRF} 360: Unbounded anti-aliased neural radiance fields},
  author={Barron, Jonathan T and Mildenhall, Ben and Verbin, Dor and Srinivasan, Pratul P and Hedman, Peter},
  booktitle={Proceedings of the IEEE/CVF Conference on Computer Vision and Pattern Recognition},
  pages={5470--5479},
  year={2022}
}

@article{stomakhinsnow,
  title={A material point method for snow simulation},
  author={Stomakhin, Alexey and Schroeder, Craig and Chai, Lawrence and Teran, Joseph and Selle, Andrew},
  journal={ACM Transactions on Graphics},
  volume={32},
  number={4},
  pages={1--10},
  year={2013}
}

@article{humlsmpm,
  title={A moving least squares material point method with displacement discontinuity and two-way rigid body coupling},
  author={Hu, Yuanming and Fang, Yu and Ge, Ziheng and Qu, Ziyin and Zhu, Yixin and Pradhana, Andre and Jiang, Chenfanfu},
  journal={ACM Transactions on Graphics},
  volume={37},
  number={4},
  pages={1--14},
  year={2018}
}
